\pdfoutput=1

\documentclass{article}

\usepackage{amsmath,amsfonts,bm}

\def\eqref#1{equation~\ref{#1}}

\def\1{\bm{1}}

\DeclareMathAlphabet{\mathsfit}{\encodingdefault}{\sfdefault}{m}{sl}
\SetMathAlphabet{\mathsfit}{bold}{\encodingdefault}{\sfdefault}{bx}{n}

\usepackage{hyperref}
\usepackage{url}
\usepackage{multirow}
\usepackage{times}
\usepackage{latexsym}
\usepackage{times}
\usepackage{float}
\usepackage{latexsym}
\usepackage{multirow}
\usepackage{url}
\usepackage{tabularx}
\usepackage{latexsym}
\usepackage{balance}
\usepackage{bm}
\usepackage{amssymb}
\usepackage{amsmath}
\usepackage{makecell}
\usepackage{multirow}
\usepackage{subfig}
\usepackage{graphicx} 
\usepackage{color}
\usepackage{colortbl}
\usepackage{textcomp}
\usepackage{CJK}
\usepackage{enumitem}
\usepackage{longtable}
\usepackage{booktabs}
\usepackage{microtype}
\usepackage{arydshln}
\usepackage{amsfonts}
\usepackage{wasysym}
\usepackage{eufrak}
\usepackage{pifont}
\usepackage[normalem]{ulem}
\usepackage[T1]{fontenc}
\usepackage{tcolorbox}
\usepackage{algorithm}
\usepackage{algorithmic}
\usepackage{xcolor,soul}
\usepackage{ellipsis}
\usepackage{wrapfig}
\tcbuselibrary{breakable}
\usepackage[normalem]{ulem}
\useunder{\uline}{\ul}{}

\definecolor{perfblue}{RGB}{64, 114, 175}
\definecolor{perfred}{RGB}{220, 60, 60}
\usepackage{hyperref}
\hypersetup{
    colorlinks=true,
    linkcolor=perfred,     
    citecolor=perfblue,     
    urlcolor=perfred,      
    pdfborder={0 0 0}   
}

 \usepackage[accepted]{icml2026}

\usepackage{amsmath}
\usepackage{amssymb}
\usepackage{mathtools}
\usepackage{amsthm}

\usepackage[capitalize,noabbrev]{cleveref}

\theoremstyle{plain}

\theoremstyle{definition}

\theoremstyle{remark}

\usepackage[textsize=tiny]{todonotes}

\icmltitlerunning{Learn to Think: Improving Multimodal Reasoning through Vision-Aware Self-Improvement Training}

\begin{document}

\twocolumn[
  \icmltitle{Learn to Think: Improving Multimodal Reasoning through \\Vision-Aware Self-Improvement Training}



  \icmlsetsymbol{equal}{*}

  \begin{icmlauthorlist}
    \icmlauthor{Qihuang~Zhong}{1}
    \icmlauthor{Liang~Ding}{2}
    \icmlauthor{Wenjie~Xuan}{1}
    \icmlauthor{Juhua~Liu}{1}
    \icmlauthor{Bo~Du}{1}
    \icmlauthor{Dacheng~Tao}{3}
  \end{icmlauthorlist}

  \icmlaffiliation{1}{School of Computer Science, National Engineering Research Center for Multimedia Software, Institute of Artificial Intelligence and Hubei Key Laboratory of Multimedia and Network Communication Engineering, Wuhan University, China}
  \icmlaffiliation{2}{The University of Sydney, Australia}
  \icmlaffiliation{3}{Nanyang Technological University, Singapore}

  \icmlcorrespondingauthor{Juhua~Liu}{liujuhua@whu.edu.cn}

  \icmlkeywords{Machine Learning, ICML}

  \vskip 0.3in
]



\printAffiliationsAndNotice{}  

\begin{abstract}
Post-training with explicit reasoning traces is common to improve the reasoning capabilities of Multimodal Large Language Models (MLLMs). However, acquiring high-quality reasoning traces is often costly and time-consuming. Hence, the \textit{self-improvement} paradigm has emerged, enabling MLLMs to self-generate reasoning traces for training without external supervision. Despite its effectiveness, we reveal two shortcomings in the self-improvement training of MLLMs: 1) data imbalance, where simple samples are over-trained, but the challenging yet crucial samples are under-trained; 2) language prior bias, where MLLMs overly rely on linguistic priors while neglecting the visual cues. To this end, we propose \texttt{\textbf{VISTA}}, a \textbf{VI}sion-aware \textbf{S}elf-improvement \textbf{T}raining framework for enhancing the multimodal re\textbf{A}soning of MLLMs. Specifically, VISTA first introduces a \textit{prefix resampling} strategy to reuse the partial correct reasoning traces for efficient data collection, and then designs a \textit{vision-aware attention score} to quantify the model’s focus on visual information. Extensive experiments show that VISTA can be applied to various post-training scenarios, \textit{i.e.}, supervised fine-tuning and preference learning, and effectively enhances the multimodal reasoning performance across various MLLMs and tasks, \textit{e.g.}, bringing up to \textbf{+13.66\%} average performance gains for Qwen2.5-VL-3B-Instruct.
\end{abstract}
\section{Introduction}
\label{sec:intro}

Inspired by the success of OpenAI o1/o3~\cite{jaech2024openai} and DeepSeek-R1~\cite{guo2025deepseek}, multimodal reasoning, which extends long chain-of-thought (CoT) capabilities to multimodal large language models (MLLMs), has recently attracted significant research attention~\cite{wang2024exploring,wang2025multimodal}. While post-training with explicit reasoning trajectories can effectively enhance the multimodal reasoning of MLLMs, their performance highly relies on high-quality intermediate reasoning traces~\citep{yangdemystifying}, which are expensive and time-consuming to obtain~\citep{pengregenesis}. Hence, recent literature~\cite{deng2025self} introduces a \textit{self-improvement} paradigm, where MLLMs iteratively improve themselves by training with their self-generated reasoning data, thus reducing the reliance on external supervision.

\begin{figure}[t]
    \centering
    \includegraphics[width=0.49\textwidth]{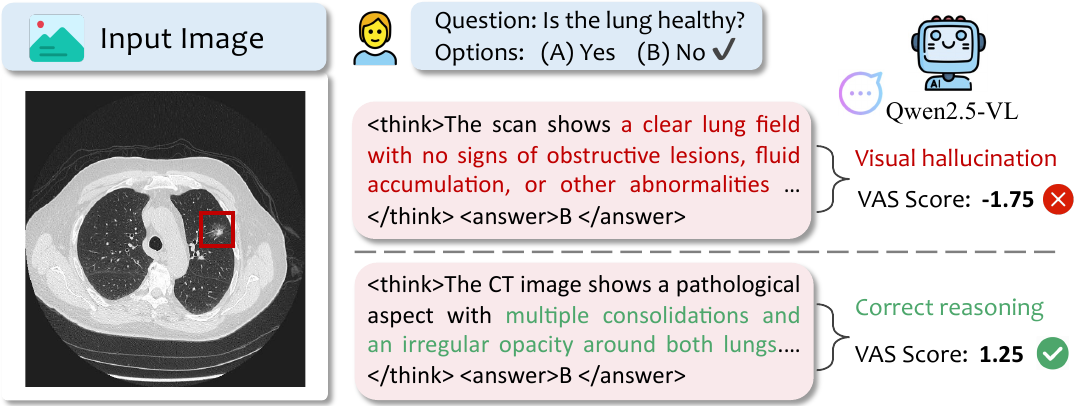}
    \caption{\textbf{Comparison of two self-generated reasoning traces}. As seen, although predicting the correct answer, MLLMs may still exhibit visual hallucinations during intermediate reasoning processes, due to over-reliance on language priors and neglect of visual cues. Encouragingly, our proposed vision-aware attention score (VAS) can accurately identify these hallucinated solutions.}
    \label{fig:figure1}
\end{figure}

In general, self-improvement training usually contains three steps: (1) data collection, where models are prompted to generate multiple reasoning solutions for each query; (2) data organization, where the ground-truth answers are used to verify the correctness of candidate solutions and the incorrect solutions will be filtered out; (3) model optimization, where models are self-trained with the selected correct solutions. 
Although self-improvement training has demonstrated remarkable performance~\cite{zelikman2022star,singhbeyond}, through a series of analyses (\S\ref{sec:preliminary}), we reveal that the self-improvement of MLLMs still suffers from two key shortcomings: \ding{182}~\textit{\textbf{data imbalance}}, \textit{i.e.}, during data collection, MLLMs can readily generate numerous correct solutions for simple queries but struggle to produce sufficient correct ones for difficult queries; \ding{183}~\textit{\textbf{language prior bias}}, \textit{i.e.}, MLLMs tend to over-rely on language priors while neglecting visual cues, leading to visual hallucinations that cannot be filtered by answer correctness alone (as illustrated in Figure~\ref{fig:figure1}). Notably, these two problems share a common root: current self-improvement methods rely solely on answer correctness as the quality signal. This single signal proves inadequate in two complementary ways: (1) in \textit{quantity}, it fails to yield sufficient correct solutions for challenging queries; (2) in \textit{quality}, it cannot distinguish visually-grounded solutions from hallucinated ones that happen to arrive at correct answers. These observations naturally give rise to two questions: \textit{1) How can we obtain sufficient correct solutions for challenging yet crucial queries? 2) How can we effectively identify and filter out undesired hallucinated solutions?}

Some prior studies also recognize these issues and attempt to address them~\cite{tong2024dart,ding2025mitigating,he2025cracking}. For \ding{182}, \citet{tong2024dart} propose to allocate more trials to difficult queries, and \citet{ding2025mitigating} attempt to use the ground-truth answers to guide the reasoning. Although effective, these methods neglect the prior failed solutions that also contain some useful information, which are wasted and inefficient. For \ding{183}, \citet{he2025cracking} design a metric to quantify the language prior bias by removing the image context and measuring the change of attention scores. Despite its remarkable performance, it requires two forward passes, leading to much computation overhead.

Different from prior works, in this paper, we propose \textbf{\texttt{VISTA}}, a \textbf{VI}sion-aware \textbf{S}elf-improvement \textbf{T}raining framework that enhances the multimodal re\textbf{A}soning of MLLMs via two novel strategies. First, to alleviate data imbalance, inspired by prior findings~\cite{ji2025first} -- \textit{errors of failed solutions often occur in the later reasoning traces}, 
we introduce a \textbf{\textit{prefix resampling}} strategy that reuses the correct prefix contexts from failed solutions for efficient data collection. In practice, it first identifies partially correct reasoning traces by locating the critical tokens that exert a significant influence on subsequent reasoning steps, and then uses these traces as prefix context to resample new solutions. Second, to mitigate the language prior bias, we design a simple-yet-effective \textbf{\textit{Vision-aware Attention Score}} (VAS) approach that leverages the model's internal attention information to quantify its focus on the visual cues during reasoning.
Intuitively, if the model allocates limited attention to the visual context, it may drift away from the visual cues, thus leading to potential visual hallucinations. Overall, by collecting more correct solutions for difficult queries and filtering the undesired solutions with lower VAS, \texttt{VISTA} can ensure the diversity and reliability of self-training data, thus bringing better reasoning performance for MLLMs.

Extensive experiments on five cutting-edge MLLMs and five popular multimodal reasoning tasks show that \texttt{VISTA} not only outperforms all counterparts by a clear margin, \textit{i.e.}, bringing up to \textbf{+13.66\%} average performance gains against the initial reasoning models, but also works well in various post-training paradigms, \textit{e.g.}, supervised fine-tuning and preference learning. More encouragingly, in-depth analyses prove that \texttt{VISTA} reduces the visual hallucinations and improves the out-of-distribution (OOD) performance effectively. To summarize, our contributions are three-fold: (1) We reveal two shortcomings of self-improvement training in MLLMs, and propose \texttt{VISTA} to address them and boost MLLMs' multimodal reasoning capabilities without extra supervision. (2) \texttt{VISTA} introduces two simple-yet-effective strategies that can be adopted to various MLLMs and post-training settings. (3) Extensive results show that \texttt{VISTA} can significantly and consistently improve MLLMs' multimodal reasoning performance and model generalization.
\section{Revisiting Self-improvement in MLLMs}

\subsection{Preliminary}
\label{sec:preliminary}

\begin{figure*}[t]
    \centering
    \includegraphics[width=\textwidth]{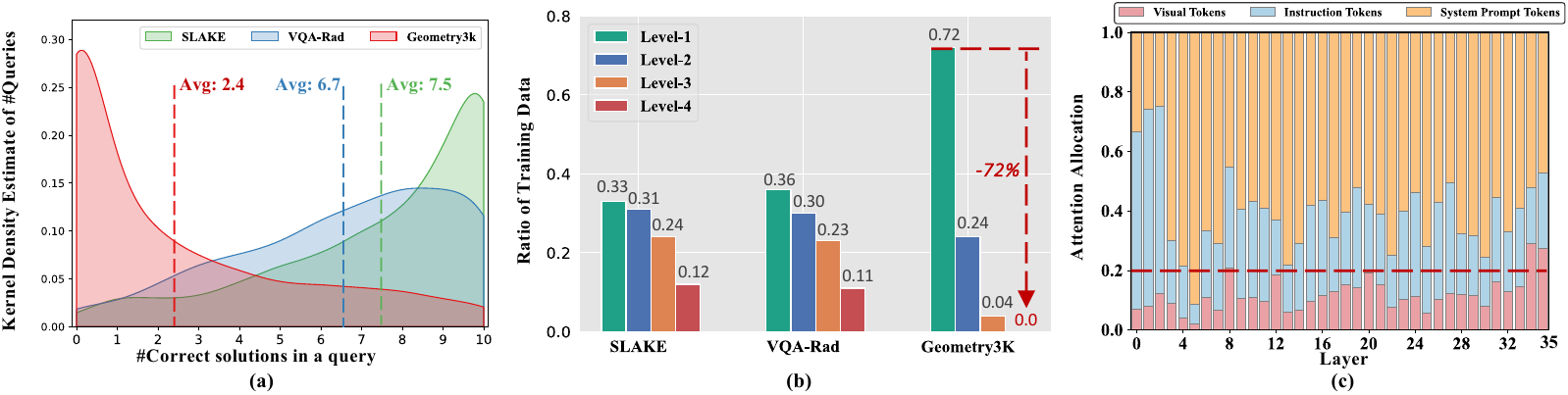}
    \caption{\textbf{(a)} Distribution of the number of correct solutions in a single query. \textbf{(b)} Distribution of self-generated training samples for different difficulty levels, where level-1 denotes the simplest and level-4 denotes the hardest. \textbf{(c)} Attention allocation between system prompts, visual, and instruction tokens across different model layers. Here, we use the Qwen2.5-VL-3B-Instruct as the base model.}
    \label{fig:preliminary_analysis}
\end{figure*}

Given a base MLLM $\mathcal{M}_{base}$ and a multimodal dataset $\mathcal{D}=\{(x_i, y_i)\}$, where $x_i = \{x^{\text{sys}}_i, x^{\text{vis}}_i, x^{\text{ins}}_i\}$ is the query and $y_i$ is corresponding ground-truth answer. $x^{\text{sys}}_i$, $x^{\text{vis}}_i$, $x^{\text{ins}}_i$ denote the system prompt, visual, and user instruction tokens, respectively. The goal of self-improvement training is to enhance the long-CoT multimodal reasoning performance of $\mathcal{M}_{base}$ by iteratively self-training using its own solutions on $\mathcal{D}$ over $T$ cycles. Let $\mathcal{M}_{t}$ denote the model at the $t$-th iteration ($t \in [1,\,T]$). The representative self-improvement training for MLLMs involves the following steps:

\textbf{Data Collection.}\quad At $t$-th iteration, for each query $x_i \in \mathcal{D}$, we enforce the previous $\mathcal{M}_{t-1}$ to generate multiple reasoning trajectories and their corresponding answers $\{(r^k_i, \hat{y}^k_i)\}^K_{k=1}$, where $K$ denotes the sampling time for each query and $r^k_i$ denotes the $k$-th trajectory. Thus, we can construct the self-generated dataset $\mathcal{D}_t=\{(x_i, r^k_i, \hat{y}^k_i)^K_{k=1}\}$.

\textbf{Data Organization.}\quad The ground-truth answer $y_i$ is used to verify the correctness of candidate solutions. According to the correctness, we can split the candidate solutions into two sets: positive set $\mathcal{D}^{p}_t=\{(x_i, r^{k_p}_i, \hat{y}^{k_p}_i)\,|\,\mathbb{I}(\hat{y}^{k_p}_i, y_i) =1 \}$ and negative set $\mathcal{D}^{n}_t=\{(x_i, r^{k_n}_i, \hat{y}^{k_n}_i)\,|\,\mathbb{I}(\hat{y}^{k_n}_i, y_i) =0 \}$.

\textbf{Model Optimization.}\quad The self-training process differs across various post-training paradigms, \textit{i.e.}, supervised fine-tuning (SFT) and preference learning. Specifically, for self-improvement SFT training, only the correct solutions from $\mathcal{D}^{p}_t$ are used to optimize the model via the standard negative log likelihood (NLL) loss function:
\begin{equation} 
\mathcal{L}_\text{SFT}=\mathop{\mathbb{E}_{\,\mathcal{D}^p_t}}\Big[-\text{log}\frac{\mathcal{M}_{\theta}(r^{k_p}_i, \hat{y}^{k_p}_i|x_i)}{|r^{k_p}_i|+|\hat{y}^{k_p}_i|}\Big],
    \label{eq:sft_loss}
\end{equation} 
where $\mathcal{M}_{\theta}$ initialized with $\mathcal{M}_{base}$ denotes the current tuned model that will become next model $\mathcal{M}_t$. Notably, following previous practice~\cite{zelikman2022star,singhbeyond}, we do not continually fine-tune $\mathcal{M}_{t-1}$ to avoid overfitting. 

For the implementation of preference learning, we use the representative Direct Preference Optimization (DPO)~\cite{rafailov2023direct}. In practice, each positive solution in $\mathcal{D}^{p}_t$ and a randomly-selected negative solution from $\mathcal{D}^{n}_t$ are paired to construct the preference training set $\mathcal{D}^{\text{pairs}}_t=\{(x_i, r^{k_p}_i, \hat{y}^{k_p}_i, r^{k_n}_i, \hat{y}^{k_n}_i)\,|\, k_p, k_n \in [1,\,K)\}$. Then, inspired by~\citet{pang2024iterative}, we leverage an enhanced DPO algorithm that combines the standard DPO and NLL loss functions to ensure the training stability:
\begin{align}
    &\mathcal{L}_\text{DPO+NLL} =  \mathcal{L}_\text{DPO} + \alpha \cdot \mathcal{L}_\text{NLL} (r^{k_p}_i, \hat{y}^{k_p}_i)  \notag \\
    &=  \mathop{\mathbb{E}_{\,\mathcal{D}^{\text{pairs}}_t}} \Big[- \log \sigma \left( f(\hat{r}^{k_p}_i, \hat{y}^{k_p}_i|x_i) - f(\hat{r}^{k_n}_i, \hat{y}^{k_n}_i|x_i)\right) \notag \\ 
    &- \alpha \cdot  \frac{\log \mathcal{M}_{\theta}(\hat{r}^{k_p}_i, \hat{y}^{k_p}_i| x_i)}{|\hat{r}^{k_p}_i| + |\hat{y}^{k_p}_i|}\Big],
\label{eq:dpo_loss}
\end{align}
where $f(\cdot|x_i)=\beta \log \frac{ \mathcal{M}_{\theta}( \cdot | x_i)}{ \mathcal{M}_{base}(\cdot | x_i)}$, $\mathcal{M}_{\theta}$ is the policy model initialized with $\mathcal{M}_{base}$, $\sigma$ is the sigmoid function. $\alpha$ and $\beta$ are coefficients that are empirically set to 0.5 and 0.1. Notably, to ensure the initial model has basic reasoning capabilities, we use some reasoning examples as few-shot demonstrations to prompt $\mathcal{M}_{base}$ to sample multiple reasoning solutions for each query. By randomly selecting one correct solution per query, we construct a seed training set and use it to fine-tune $\mathcal{M}_{base}$ for obtaining the initial $\mathcal{M}_{0}$.

\subsection{Empirical Analyses}
\label{sec:preliminary_analysis}

\textbf{Settings.}\quad Here, we use the Qwen2.5-VL-3B-Instruct~\cite{bai2025qwen2} model and several popular multimodal reasoning benchmarks, \textit{i.e.}, SLAKE~\cite{liu2021slake}, VQA-Rad~\cite{lau2018dataset} and Geometry3K~\cite{lu2021inter}, as a testbed. During the implementation of self-improvement training, the sampling time $K$ is set to 10, and the number of iterations $T$ is set to 1 for faster experimental validation. 

\textbf{Findings.}\quad Through a series of analyses on the self-generated samples, we empirically found two issues:

\ding{182} \textbf{Data Imbalance}: First, we illustrate the distribution of the number of self-generated correct solutions per query in Figure~\ref{fig:preliminary_analysis} (\textbf{a}). As seen, the number of correct solutions is highly related to task difficulty, where MLLMs can produce numerous correct solutions for simpler tasks (\textit{i.e.}, SLAKE) but struggle to generate sufficient correct ones for harder tasks (\textit{i.e.}, Geometry3K). Specifically, for Geometry3K, there are more than 40\% queries without any correct solutions. Unfortunately, these challenging queries are proven to be more crucial for further training~\citep{liumakes}. To have a closer look, we evenly split all queries into four levels based on the number of correct solutions per query, and illustrate the distribution of self-generated training samples at different levels in Figure~\ref{fig:preliminary_analysis} (\textbf{b}). Obviously, most self-generated samples are simple, whereas the challenging yet crucial samples are scarce, especially in harder tasks.

 \begin{figure*}[t]
     \centering
     \includegraphics[width=1\linewidth]{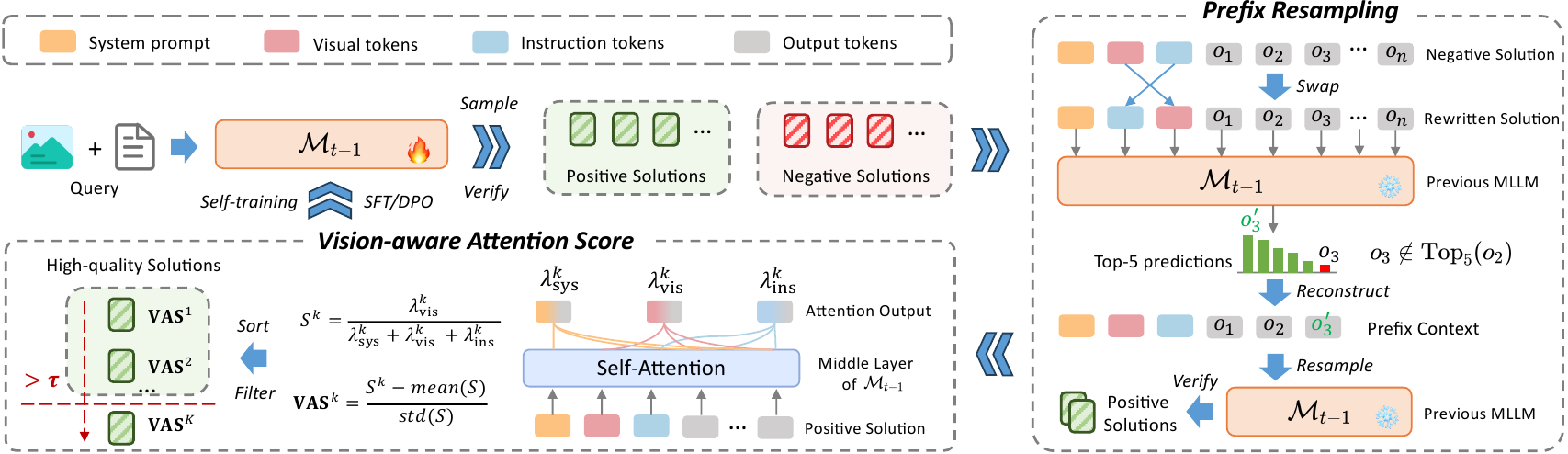}
     \caption{\textbf{Overview of our \texttt{VISTA} framework}, which consists of two simple-yet-effective strategies: 1) \textbf{\textit{prefix resampling}}, aiming to collect more accurate solutions for difficult queries; 2) \textbf{\textit{vision-aware attention score}}, aiming to filter out undesired hallucinated solutions.
     }
     \label{fig:method}
 \end{figure*}

\ding{183} \textbf{Language Prior Bias}: By analyzing the candidate solutions in $\mathcal{D}^p_t$, we found that, while some solutions make the correct predictions, they may exhibit hallucination issues in the reasoning traces, \textit{e.g.}, 
describing some things that do not exist in the image, or exhibiting a conflict between the reasoning chain and its final answer.
Some examples are provided in Figure~\ref{fig:hallucinated_data_case} of Appendix~\ref{sec:vas_analysis}. To investigate this problem, inspired by prior works~\cite{leng2024mitigating,liu2025more,he2025cracking} related to hallucination alleviation, we leverage the model's internal attention information to directly probe the model's behavior. Specifically, we follow~\citet{liu2025more} and compare the attention distributions of the first output token over visual, instruction, and system prompt tokens across all layers. In practice, we randomly sample 100 data points from the training set of VQA-Rad, and illustrate the average attention distributions in Figure~\ref{fig:preliminary_analysis} (\textbf{c}). It can be found that, although visual tokens account for the largest proportion in the context, they generally receive the lowest attention scores (less than 20\% in most layers) across all layers. Conversely, the model tends to focus on the linguistic information of instruction and system prompt tokens. We conjecture that visual hallucinations in MLLMs may originate from such a language prior bias.



\section{Methodology}
\label{sec:method}


\textbf{Motivation of \texttt{VISTA}.}\quad  To address the above problems, we propose \texttt{VISTA} that harnesses self-improvement in MLLMs via two simple-yet-effective approaches. First, for \ding{182}, we build upon the insight~\cite{ji2025first,yang2025dynamic} that \textit{errors of failed solutions often occur in the later reasoning traces, while early reasoning steps are usually correct and informative}. Instead of discarding these valuable prior failed solutions, we introduce \textit{\textbf{prefix resampling}} to accurately locate the partially correct prefix and reuse it to achieve efficient data collection. Second, for \ding{183}, motivated by the empirical analysis in \S\ref{sec:preliminary_analysis}, we recognize that the internal information of MLLMs' attention heads can reflect the language prior bias effectively. Hence, we further introduce \textbf{\textit{Vision-aware Attention Score}} (VAS), an efficient metric that measures the attention allocation over visual tokens in reasoning trajectories to quantify the model's focus on visual cues. Figure~\ref{fig:method} illustrates the overview of \texttt{VISTA}, and the pseudo-code of \texttt{VISTA} is shown in Algorithm~\ref{alg:vista}. 

\textbf{Prefix Resampling.}\quad  The core of prefix resampling is to locate the correct prefix. Motivated by the concept of \textit{critical tokens} -- some output tokens exert significant influence on later reasoning steps~\cite{lincritical}, we attempt to make full use of MLLMs' self-calibration abilities to locate these critical tokens from prior solutions without relying on ground-truth answers and extra models. Specifically, for each sample in $\mathcal{D}^{n}_t$, we construct a new query by swapping the position\footnote{The effectiveness of the swap method, along with further empirical analysis, is provided in Appendix~\ref{sec:resampling_analysis}.} of the image, \textit{e.g.}, from ``$x^{\text{vis}}_i + x^{\text{ins}}_i$'' to ``$x^{\text{ins}}_i + x^{\text{vis}}_i$'', and concatenate the new query with prior reasoning traces, \textit{i.e.}, ``$x^{\text{sys}}_i + x^{\text{ins}}_i + x^{\text{vis}}_i + r^{k_n}_i$''. The reconstructed sample is fed into $\mathcal{M}_{t-1}$ to obtain the Top-5 token predictions\footnote{The effect of the Top-$k$ setting is analyzed in Appendix~\ref{sec:resampling_analysis}.} at the next position for each reasoning token, denoted as ${\text{Top}_5(o_n)}$, where $o_n$ means the $n-$th token in the reasoning trajectory. The first token that does not match the expected Top-5 tokens, \textit{i.e.}, $o_n \notin \text{Top}_5(o_{n-1})$, is regarded as the critical token. Then, we replace the $o_n$ with the new Top-1 predicted token $o^{'}_n=\text{Top}_1(o_{n-1})$ and truncate the subsequent reasoning steps. Lastly, we concatenate the original query with truncated reasoning traces as a prefix and resample the subsequent reasoning steps for $J$ times to collect the correct solutions, which are merged into $\mathcal{D}^{p}_t$.

\textbf{Vision-aware Attention Score.}\quad  To alleviate the language prior bias, we leverage the MLLMs' internal attention information to quantify the model's focus on visual cues. 
Specifically, for each correct solution $(x^{\text{sys}}_i, x^{\text{vis}}_i, x^{\text{ins}}_i, r^k_i, \hat{y}^k_i)$ in $\mathcal{D}^p_t$, we extract its attention output $\mathbf{A}^k_i$ from an intermediate layer of $\mathcal{M}_{t-1}$. Since the middle layers play a crucial role in processing visual information~\cite{jiang2025devils}, we select the middle layer of $\mathcal{M}_{t-1}$ in this work (further analyses are provided in Appendix~\ref{sec:vas_analysis}).
Then, we measure the attention allocation assigned by the output reasoning tokens to different contexts:
\begin{equation}
\begin{aligned}
	\lambda^k_{\text{sys}} &= {\textstyle \sum^{(r^k_i, \hat{y}^k_i)}_{o}}  {\textstyle \sum^{x^{\text{sys}}_i}_{s}}\mathbf{A}^k_i(o, s), \\
	\lambda^k_{\text{vis}} &= {\textstyle \sum^{(r^k_i, \hat{y}^k_i)}_{o}}  {\textstyle \sum^{x^{\text{vis}}_i}_{s}}\mathbf{A}^k_i(o, s), \\
	\lambda^k_{\text{ins}} &= {\textstyle \sum^{(r^k_i, \hat{y}^k_i)}_{o}}  {\textstyle \sum^{x^{\text{ins}}_i}_{s}}\mathbf{A}^k_i(o, s), 
\label{eq:attention}
\end{aligned}
\end{equation}
where $\lambda^k_{\text{sys}}$, $\lambda^k_{\text{vis}}$, and $\lambda^k_{\text{ins}}$ denote the attention allocation over system tokens, visual tokens, and instruction tokens, respectively. Based on it, we leverage the degree of $\lambda^k_{\text{vis}}$ to quantify the focus of MLLMs on visual cues, and regularize different scores using a standard normalization:
\begin{equation}
		S^k_i =  \frac{\lambda^k_{\text{vis}}}{\lambda^k_{\text{sys}}+\lambda^k_{\text{vis}}+\lambda^k_{\text{ins}}}, \text{\textbf{VAS}}^k_i = \frac{S^k_i-mean({S_i})}{std({S_i})},
\label{eq:vas}
\end{equation}
where $\text{\textbf{VAS}}^k_i$ refers to our proposed VAS score. The undesired solutions with scores below the threshold $\tau$ are filtered, and the remaining high-quality correct solutions are used to perform the iterative self-improvement training.

Intuitively, by paraphrasing the query and verifying whether the prior generated token matches the expected token predictions, we can leverage MLLMs' self-calibration capabilities to identify uncertain tokens that may lead to deviating from the correct reasoning trajectory. Moreover, by replacing these uncertain tokens with more reliable ones, we can efficiently resample more accurate solutions for difficult queries. Furthermore, we use our VAS metric to filter out the undesired solutions that overly rely on language priors. By doing so, \texttt{VISTA} can construct a more diverse and reliable reasoning dataset, thus leading to better performance.

\begin{table*}[h]
\centering
\caption{\textbf{Performance comparison between Qwen2.5-VL family models} using different training methods on medical and mathematical multimodal reasoning benchmarks. ``|Train|'' means the average number of training samples among all models and tasks. ``Overall'' denotes the average accuracy, while the subscript results denote the performance gains against the SFT-Seed. Best results are in \textbf{bold}.}
\label{tab:main_result1}
\setlength{\tabcolsep}{9pt}
\resizebox{1\textwidth}{!}{
\begin{tabular}{lcccllccll}
\toprule
\multirow{2}{*}{\textbf{Methods}} & \textbf{|Train|} & \multicolumn{4}{c}{\textbf{Qwen2.5-VL-3B-Instruct}} & \multicolumn{4}{c}{\textbf{Qwen2.5-VL-7B-Instruct}} \\ \cmidrule(lr){3-6} \cmidrule(lr){7-10}
 &{\small \textbf{Avg.}} &{\small \textbf{SLAKE}} &{\small \textbf{VQA-Rad}} &{\small \textbf{Geo3K}} &{\small \textbf{Overall ($\Delta$)}} &{\small \textbf{SLAKE}} &{\small \textbf{VQA-Rad}} &{\small \textbf{Geo3K}} &{\small \textbf{Overall ($\Delta$)}} \\ \midrule
SFT-Seed & 1.0K & 67.04 & 64.14 & 25.46 & 52.21 & 79.15 & 70.52 & 36.94 & 62.20 \\
SFT-Oracle & 1.3K  & 81.97 & 68.13 & 37.94 & 62.68\textcolor{green!70!black}{{$_{\uparrow 10.47}$}} & 82.25 & 70.12 & 41.43 & 64.60\textcolor{green!70!black}{{$_{\uparrow 2.40}$}} \\ \midrule
RFT & 25.5K & 79.15 & 70.12 & 28.79 & 59.35\textcolor{green!70!black}{{$_{\uparrow 7.14}$}} & 82.82 & 74.90 & 40.27 & 66.00\textcolor{green!70!black}{{$_{\uparrow 3.80}$}} \\ \hdashline
\rowcolor{gray!20} \multicolumn{10}{l}{STaR} \\
\quad Iteration 1 & 0.9K  & 73.80 & 66.13 & 32.11 & 57.35\textcolor{green!70!black}{{$_{\uparrow 5.14}$}} & 80.00 & 71.31 & 37.44 & 62.92\textcolor{green!70!black}{{$_{\uparrow 0.72}$}} \\
\quad Iteration 2 & 1.0K  & 76.90 & 65.34 & 31.95 & 58.06\textcolor{green!70!black}{{$_{\uparrow 5.85}$}} & 80.00 & 72.11 & 37.94 & 63.35\textcolor{green!70!black}{{$_{\uparrow 1.15}$}} \\
\quad Iteration 3 & 1.1K & 80.28 & 69.32 & 30.78 & 60.13\textcolor{green!70!black}{{$_{\uparrow 7.92}$}} & 84.23 & 69.72 & 37.10 & 63.68\textcolor{green!70!black}{{$_{\uparrow 1.48}$}} \\ \hdashline
\rowcolor{gray!20} \multicolumn{10}{l}{ReST$^{EM}$} \\
\quad Iteration 1 & 8.5K & 76.06 & 69.72 & 29.95 & 58.58\textcolor{green!70!black}{{$_{\uparrow 6.37}$}} & 82.82 & 73.31 & 35.77 & 63.97\textcolor{green!70!black}{{$_{\uparrow 1.77}$}} \\
\quad Iteration 2 & 9.9K & 79.38 & 71.31 & 31.45 & 60.71\textcolor{green!70!black}{{$_{\uparrow 8.50}$}} & 84.79 & 74.49 & 39.77 & 66.35\textcolor{green!70!black}{{$_{\uparrow 4.15}$}} \\
\quad Iteration 3 & 10.7K & 81.69 & 73.71 & 32.28 & 62.56\textcolor{green!70!black}{{$_{\uparrow 10.35}$}} & 84.51 & 75.30 & 38.94 & 66.25\textcolor{green!70!black}{{$_{\uparrow 4.05}$}} \\ \hdashline
\rowcolor{gray!20} \multicolumn{10}{l}{R3V} \\
\quad Iteration 1 & 18.5K  & 78.21 & 65.74 & 29.62 & 57.89\textcolor{green!70!black}{{$_{\uparrow 5.68}$}}  & 82.82 & 75.30 & 36.44 & 64.85\textcolor{green!70!black}{{$_{\uparrow 2.65}$}} \\
\quad Iteration 2 & 18.8K  & 82.82 & 74.10 & 32.61  & 63.18\textcolor{green!70!black}{{$_{\uparrow 10.97}$}} & 86.48 & 75.30 & 38.44 & 66.74\textcolor{green!70!black}{{$_{\uparrow 4.54}$}}  \\
\quad Iteration 3 & 18.7K & 81.41 & 69.32 & 32.78 & 61.17\textcolor{green!70!black}{{$_{\uparrow 8.96}$}} & 87.32 & 75.70 & 37.27 & 66.76\textcolor{green!70!black}{{$_{\uparrow 4.56}$}} \\ \hdashline
\rowcolor[RGB]{233,246,255} \multicolumn{10}{l}{\textbf{\texttt{VISTA-SFT} (Ours)}} \\
\quad Iteration 1 & 7.7K & 80.85 & 74.10 & 32.28 & 62.41\textcolor{green!70!black}{{$_{\uparrow 10.20}$}} & 83.66 & 74.90 & 40.77 & 66.44\textcolor{green!70!black}{{$_{\uparrow 4.24}$}} \\
\quad Iteration 2 & 8.1K & 83.38 & 72.51 & 34.28 & 63.39\textcolor{green!70!black}{{$_{\uparrow 11.18}$}} & 84.79 & 75.30 & 41.26 & 67.12\textcolor{green!70!black}{{$_{\uparrow 4.92}$}} \\
\quad Iteration 3 & 8.3K & \textbf{84.23} & \textbf{76.10} & \textbf{37.27} & \textbf{65.87\textcolor{green!70!black}{{$_{\uparrow \textbf{13.66}}$}}} & \textbf{87.89} & \textbf{77.29} & \textbf{41.43} & \textbf{68.87}\textcolor{green!70!black}{{$_{\uparrow \textbf{6.67}}$}} \\ 
\bottomrule
\end{tabular}
}
\end{table*}

\section{Experiments}
\label{sec:experiments}
\subsection{Experimental Setup}
\textbf{Tasks and Datasets.}\quad We mainly assess the effectiveness of \texttt{VISTA} on multimodal medical reasoning and mathematical reasoning tasks, \textit{i.e.}, using the SLAKE~\cite{liu2021slake} and VQA-Rad~\cite{lau2018dataset} for medical reasoning, while using the Geometry3K~\cite{lu2021inter} (Geo3K for short) for mathematical reasoning. Moreover, to verify the universality of \texttt{VISTA}, we also evaluate on two popular multimodal reasoning benchmarks, including the ScienceQA~\cite{lu2022learn} and ChartQA~\cite{masry2022chartqa}. For each task, we post-train MLLMs on the training set and evaluate them on the corresponding test set by using the zero-shot accuracy as the metric. Following DeepSeek-R1~\cite{guo2025deepseek}, the reasoning process and answer are enclosed with \texttt{<think></think>} and \texttt{<answer></answer>}. More details of the used datasets are shown in Appendix~\ref{sec:dataset_details}.

\textbf{Training Details.}\quad We conduct main experiments using several cutting-edge and powerful MLLMs, including \texttt{Qwen2.5-VL-3B/7B-Instruct}~\cite{bai2025qwen2}, \texttt{Qwen3-VL-2B-Instruct}~\cite{Qwen3-VL}, \texttt{InternVL3-2B/8B}~\cite{zhu2025internvl3} models. During the implementation of our \texttt{VISTA}, the sampling times $K$ and $J$ are set to 10 and 3, respectively. The sampling temperature is 1.0, and the maximum output length is 2,048. The threshold $\tau$ is set to -0.5. For the SFT of Qwen2.5-VL models, we set the self-improvement iteration $T$ to 3. Notably, due to the limited computational resources, for the other models, we only perform the self-improvement training for one iteration, \textit{i.e.}, $T=1$. During model inference, greedy decoding with a maximum output length of 2,048 is used for reproducibility. All models are post-trained for 3 epochs on 8 NVIDIA A800 (80GB) GPUs. More training and inference details are provided in Appendix~\ref{sec:training_details}.

\begin{table*}[t]
\centering
\caption{\textbf{Performance comparison on more MLLMs} tuned with different self-improvement SFT methods on several reasoning benchmarks. In this experiment, we perform the self-improvement training for one iteration. Best results are in \textbf{bold}.}
\label{tab:main_result2}
\setlength{\tabcolsep}{8pt}
\resizebox{1\textwidth}{!}{
\begin{tabular}{lcccccccccl}
\toprule
\multirow{2}{*}{\textbf{Methods}} & \multicolumn{3}{c}{\textbf{Qwen3-VL-2B-Instruct}} & \multicolumn{3}{c}{\textbf{InternVL3-2B}} & \multicolumn{3}{c}{\textbf{InternVL3-8B}} & \multicolumn{1}{c}{\textbf{Overall}} \\ \cmidrule(lr){2-4} \cmidrule(lr){5-7} \cmidrule(lr){8-10}
 & {\small \textbf{SLAKE}} &{\small \textbf{VQA-Rad}} &{\small \textbf{Geo3K}} & {\small \textbf{SLAKE}} &{\small \textbf{VQA-Rad}} &{\small \textbf{Geo3K}} & {\small \textbf{SLAKE}} &{\small \textbf{VQA-Rad}} &{\small \textbf{Geo3K}} & \multicolumn{1}{c}{($\Delta$)} \\ \midrule
SFT-Seed & 69.86 & 68.92 & 32.61 & 71.54 & 58.96 & 26.12 & 84.51 & 75.70 & 41.60 &58.87 \\
SFT-Oracle & 80.28 & 68.92 & 37.43 & 81.97 & 72.11 & 32.78 & 86.10 & 76.49 & 45.59 & 64.63\textcolor{green!70!black}{{$_{\uparrow 5.76}$}}\\ \midrule
STaR & 75.21 & 65.34 & 34.28 & 71.55 & 66.53 & 29.78 & 85.92 & 74.50 & 41.93  &60.56\textcolor{green!70!black}{{$_{\uparrow 1.69}$}} \\
STaR+ & 78.87 & 69.32 & 33.78 & 81.41 & 68.92 & 31.28 & 86.76 & 72.51 & 43.43 & 62.92\textcolor{green!70!black}{{$_{\uparrow 4.05}$}} \\
R3V & 81.69 & 66.14 & 35.77 & 80.00 & 72.31 & 29.12 & 87.04 & 72.51 & 43.76 &63.15\textcolor{green!70!black}{{$_{\uparrow 4.28}$}} \\
\rowcolor[RGB]{233,246,255} \textbf{\texttt{VISTA-SFT}}& \bf 81.97 & \bf 71.31 & \bf 36.27 & \bf 82.54 & \bf 72.51 & \bf 31.45 & \bf 87.61 & \bf 78.49 & \bf 45.59 & \bf 65.30\textcolor{green!70!black}{{$_{\uparrow 6.43}$}} \\
\bottomrule
\end{tabular}
}
\end{table*}

\begin{table*}[t]
\centering
\begin{minipage}{0.48\linewidth}
  \centering
  \caption{Performance comparison between different tuned Qwen2.5-VL models on \textbf{more reasoning benchmarks}.}
  \label{tab:more_tasks}
  \resizebox{1\linewidth}{!}{
  \begin{tabular}{lcccc}
    \toprule
    \multirow{2}{*}{\bf Methods} & \multicolumn{2}{c}{\bf Qwen2.5-VL-3B} & \multicolumn{2}{c}{\bf Qwen2.5-VL-7B} \\ \cmidrule(lr){2-3} \cmidrule(lr){4-5}
     &\bf \small ScienceQA &\bf \small ChartQA &\bf \small ScienceQA &\bf \small ChartQA \\ \midrule
    SFT-Seed & 74.66 & 60.82 & 81.55 & 72.32 \\
    SFT-Oracle &82.00 &65.98	&85.49	&75.34\\ \midrule
    STaR & 77.49 & 61.18 & 84.39 & 72.46 \\
    STaR+ & 78.60 & 61.89 & 85.73 & 72.81 \\
    R3V & 80.81 & 61.31 & 86.83 & 73.10 \\
    \rowcolor[RGB]{233,246,255} \bf \texttt{VISTA-SFT}&\bf 82.29 &\bf 62.09 &\bf 87.58 &\bf 73.68 \\
    \bottomrule
  \end{tabular}
  }
\end{minipage}
\hspace{\fill}
\begin{minipage}{0.48\linewidth}
  \centering
   \caption{Performance comparison between tuned Qwen2.5-VL-3B-Instruct models using \textbf{different post-training algorithms}.}
  \label{tab:post_training}
  \resizebox{1\linewidth}{!}{
  \begin{tabular}{lccll}
    \toprule
    \textbf{Methods} & {\small \textbf{SLAKE}} &{\small \textbf{VQA-Rad}} &{\small \textbf{Geo3K}} &{\small \textbf{Overall ($\Delta$)}} \\ \midrule
    SFT-Seed & 67.04 & 64.14 & 25.46 & 52.21 \\ \midrule
    \rowcolor{gray!20} \multicolumn{5}{l}{\textit{(a) DPO Training}} \\
    IRPO & 86.20 & 74.10 & 29.11 & 63.14\textcolor{green!70!black}{{$_{\uparrow 10.93}$}} \\
    \rowcolor[RGB]{233,246,255} \bf \texttt{VISTA-DPO} &\bf 89.01 &\bf 76.89 &\bf 29.45 &\bf 65.12\textcolor{green!70!black}{{$_{\uparrow 12.91}$}} \\ \midrule
    \rowcolor{gray!20} \multicolumn{5}{l}{\textit{(b) GRPO Training, initialized by SFT models}} \\
    ReST$^{EM}$-GRPO & 84.51 & 78.09 & 32.45 & 65.02\textcolor{green!70!black}{{$_{\uparrow 12.81}$}} \\
    \rowcolor[RGB]{233,246,255} \bf \texttt{VISTA-GRPO} &\bf 85.35 &\bf 79.28 &\bf 32.61 &\bf 65.75\textcolor{green!70!black}{{$_{\uparrow 13.54}$}} \\
    \bottomrule
  \end{tabular}
  }
\end{minipage}
\end{table*}

\textbf{Baselines.}\quad We compare \texttt{VISTA} with several representative and cutting-edge self-improvement training methods:

\begin{itemize}[leftmargin=10pt, itemindent=0pt]
	\item \textbf{SFT-Seed}: Standard fine-tuning $\mathcal{M}_{base}$ on the self-generated seed data $\mathcal{S}$ to obtain the initial model $\mathcal{M}_0$.
	\item \textbf{SFT-Oracle}: Standard fine-tuning $\mathcal{M}_{base}$ on $\mathcal{D}$ with reasoning trajectories distilled from a third-party MLLM, \textit{i.e.}, Qwen3-VL-32B-Instruct, which can be considered as the upper bound of self-improvement SFT training.
	\item \textbf{STaR}~\citep{zelikman2022star}: Sampling a solution using greedy decoding for each query in $\mathcal{D}$, where the correct solutions are used to iteratively fine-tune $\mathcal{M}_{base}$.
	\item \textbf{ReST$^{EM}$}~\citep{singhbeyond}: Extending STaR by sampling $K$ solutions for each query in $\mathcal{D}$, where all correct solutions are used for iterative SFT training.
	\item \textbf{RFT}~\citep{yuan2023scaling}: Similar to ReST$^{EM}$ but not iterative. To maintain consistent training budgets, we sample $T \times K$ candidate solutions for each query in $\mathcal{D}$. 
	\item \textbf{R3V}~\citep{cheng2025vision}: Extending ReST$^{EM}$ by additionally using the incorrect solutions to construct self-refine and self-select data for iterative SFT training.
	\item \textbf{IRPO}~\citep{pang2024iterative}: Sampling $K$ solutions for each query, where correct and incorrect solutions are paired to construct the preference data for iterative DPO training.
\end{itemize}

For a fair comparison, we keep a fixed data synthesis budget for all baselines, except STaR. Considering our goal of proposing a self-improvement training method, in this part, we do not compare \texttt{VISTA} with inference-time methods, \textit{e.g.}, self-consistency~\cite{wangself}. More comparisons with inference-time methods are shown in Appendix~\ref{sec:self_consistency}.

\begin{figure*}[t]
    \centering
    \includegraphics[width=\textwidth]{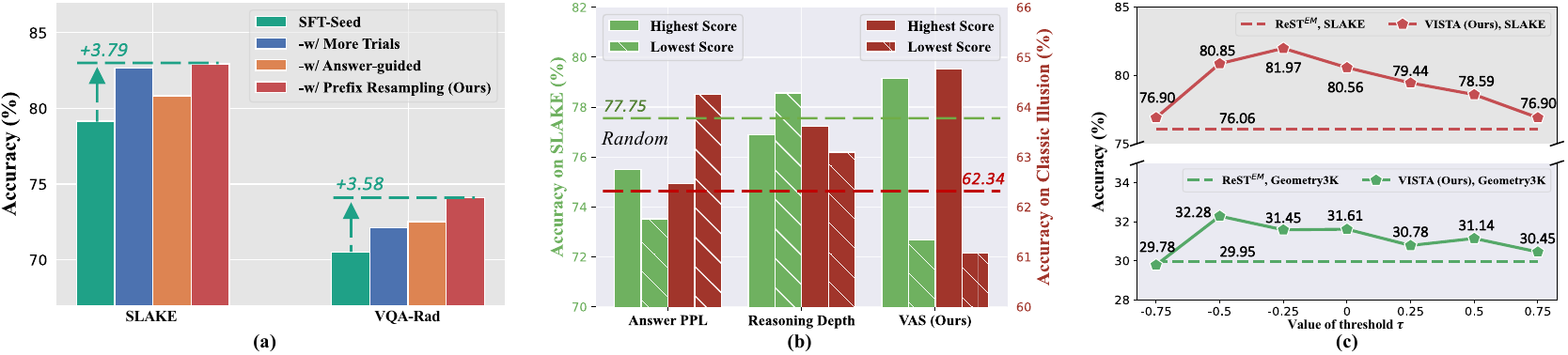}
    \caption{\textbf{(a)} Performance comparison of tuned Qwen2.5-VL-7B models using different data collection methods. \textbf{(b)} Performance comparison of tuned Qwen2.5-VL-3B models using different data selection metrics on SLAKE. \textbf{(c)} Parameter analysis of threshold $\tau$ in \texttt{VISTA} on Qwen2.5-VL-3B models. Notably, in these experiments, we perform the self-improvement SFT training for one iteration.}
    \label{fig:stage_analysis_all_results}
\end{figure*}

\subsection{Main Results}

\textbf{\texttt{VISTA} outperforms the other baseline methods by a clear margin.}\quad Table~\ref{tab:main_result1} reports the comparative results (\%) of Qwen2.5-VL family models on medical and mathematical reasoning tasks. As seen, compared to the SFT-Seed baseline, almost all self-improvement methods bring some performance improvements, proving the effectiveness of the self-improvement training paradigm. Among which, our \texttt{VISTA} outperforms all counterparts and leads to the highest performance gains, \textit{i.e.}, \textbf{+13.66\%} and \textbf{+6.67\%} average scores for 3B and 7B Qwen2.5-VL models, respectively. It is noteworthy that some baseline methods suffer from model collapse, where the model's performance degrades due to iterative self-training on model-generated data~\cite{bertrandstability}, especially in the harder Geometry3K task. We attribute it to the data imbalance problem, as over-training on the simple queries could lead to over-fitting. Conversely, by sampling more accurate solutions for challenging queries, our \texttt{VISTA} can alleviate this problem and continue to bring better performance gains, proving its effectiveness.

\textbf{\texttt{VISTA} brings consistent and significant performance gains across various models and tasks.}\quad In addition to evaluating Qwen2.5-VL models on medical and mathematical tasks, we also adopt our \texttt{VISTA} method to more models and tasks. Specifically, Table~\ref{tab:main_result2} reports the comparative results on the other MLLMs, and Table~\ref{tab:more_tasks} presents the results of different tuned Qwen2.5-VL models on more tasks. Notably, in both experiments, we only perform the self-improvement SFT training for one iteration. On the one hand, from Table~\ref{tab:main_result2}, we see that our \texttt{VISTA} still works in all these models, and achieves the best performance (even outperforms the SFT-Oracle) among all settings. Specifically, compared to SFT-Seed, \texttt{VISTA} brings \textbf{+6.43\%} average performance gains. On the other hand, as shown in Table~\ref{tab:more_tasks}, our \texttt{VISTA} consistently outperforms the other counterparts by a clear margin in both ScienceQA and ChartQA tasks. These results can prove the generality of \texttt{VISTA}.

\textbf{\texttt{VISTA} can be applied to various post-training paradigms.}\quad In the above experiments, we evaluate our \texttt{VISTA} in the self-improvement SFT training setting. Here, we conduct more experiments to verify whether \texttt{VISTA} works well in the other post-training paradigms, \textit{i.e.}, DPO~\cite{rafailov2023direct} and GRPO~\cite{shao2024deepseekmath} training. For DPO training, we compare \texttt{VISTA} with the IRPO~\citep{pang2024iterative} baseline, both of which aim to optimize the base model $\mathcal{M}_{base}$ with the self-generated preference data. As for GRPO training, we initialize the policy model with the cold-start models tuned with different methods. The ReST$^{EM}$ is used as the baseline in this setting. In both experiments, we use the Qwen2.5-VL-3B-Instruct as the base model and perform the self-improvement post-training for one iteration. The comparative results are shown in Table~\ref{tab:post_training}, from which we observe that: 1) Compared to SFT training, both DPO and GRPO training methods lead to much better performance, indicating the superiority of reinforcement learning in the reasoning MLLM field. 2) Our \texttt{VISTA} continues to surpass the baseline methods in both post-training paradigms, \textit{e.g.}, bringing +1.98\% average gains against the powerful IRPO during DPO training.

\subsection{More Analyses}

\textbf{Impact of Prefix Resampling Strategy.}\quad We first validate the effect of our \textit{prefix resampling} strategy by comparing it with two representative baselines: 1) ``-w/ More Trials'' that allocates more trials to difficult queries~\cite{tong2024dart}; 2) ``-w/ Answer-guided'' that leverages the ground-truth answers to guide the reasoning~\cite{ding2025mitigating}.
Notably, for a fair comparison, we ensure that the amount of sampled training data among all methods is comparable, and do not use the subsequent VAS method. The comparative results on Qwen2.5-VL-7B models are shown in Figure~\ref{fig:stage_analysis_all_results} \textbf{(a)}.
From these results, we can see that all methods consistently outperform the SFT-Seed baseline, confirming the importance of mitigating data imbalance. Among these methods, our prefix resampling strategy performs best in both reasoning tasks. We conjecture that the partially correct prefix contexts are more informative and reduce the difficulty of sampling more accurate solutions for challenging queries, thus alleviating the data imbalance problem effectively. More in-depth analyses of prefix resampling are shown in Appendix~\ref{sec:resampling_analysis}.

\textbf{Impact of Vision-aware Attention Score.}\quad In this part, we investigate the effect of our \textit{vision-aware score} that aims to identify and filter out the undesired hallucinated solutions. For comparison, we use two representative counterparts as the baselines: 1) ``\textit{Answer PPL}'' that employs the perplexity of solutions to quantify the model's uncertainty during the reasoning~\cite{caoinstruction}, where solutions with smaller PPL are referred to better; 2) ``\textit{Reasoning Depth}'' that leverages the JSD divergence between reasoning tokens' hidden states at the final model layer and middle layer to quantify the reasoning depth~\cite{chuang2023dola}, where solutions with larger JSD scores are referred to better. In practice, after obtaining the self-generated candidate solutions via our prefix resampling strategy, we construct two self-training sets by selecting one solution for each query with the highest and lowest score across different metrics. Then, Qwen2.5-VL-3B-Instruct is fine-tuned on these two sets, respectively. In addition to the SLAKE task, all models are also evaluated on a sub-task of IllusionBench~\cite{zhang2025illusionbench}, ``Classic Illusion'', which is a cutting-edge multimodal hallucination detection benchmark. For reference, we also construct a training set by randomly selecting one correct solution for each query, and use it to fine-tune the model.
The comparative results are illustrated in Figure~\ref{fig:stage_analysis_all_results} \textbf{(b)}.

From the results on SLAKE, we can see that ``\textit{Answer PPL}'' and ``\textit{Reasoning Depth}'' metrics perform poorly in the self-improvement training of MLLMs, as they either cannot adequately meet the needs of model training (samples with high scores perform worse) or even underperform the random baseline. Conversely, our vision-aware attention score can effectively reflect the quality of self-generated data, as training on solutions with the lowest scores leads to catastrophic performance degradation. This underscores the importance to alleviate the language prior bias problem, and to encourage the MLLMs to focus more on visual cues. This conclusion can also be proven by the results on the Classic Illusion set, as training on undesired solutions identified by our metric leads to more hallucinations. These results demonstrate the effectiveness of our metric. Additionally, we provide more analyses of our VAS metric in Appendix~\ref{sec:vas_analysis}.

\textbf{Parameter Analysis.}\quad The threshold $\tau$, used to filter the undesired solutions, is an important hyperparameter in our \texttt{VISTA} framework. In this part, we analyze its effect by evaluating the \texttt{VISTA} performance with different $\tau$ values, ranging from -0.75 to 0.75. The Qwen2.5-VL-3B-Instruct is used as the base model and evaluated on SLAKE and Geometry3K tasks. For reference, we compare our \texttt{VISTA} with the ReST$^{EM}$ baseline, and present the results in Figure~\ref{fig:stage_analysis_all_results} \textbf{(c)}. As seen, compared to the baseline, \texttt{VISTA} with various $\tau$ can achieve better performance, proving that \texttt{VISTA} is not very sensitive to the threshold. More specifically, too large $\tau$ (\textit{e.g.}, 0.75) would lead to performance degradation, as many useful solutions might be filtered. Overall, in the case of $\tau=-0.5$, \texttt{VISTA} achieves the optimal performance, thus leaving it as our default setting in this work.

\begin{figure}[t]
    \centering
    \includegraphics[width=0.75\linewidth]{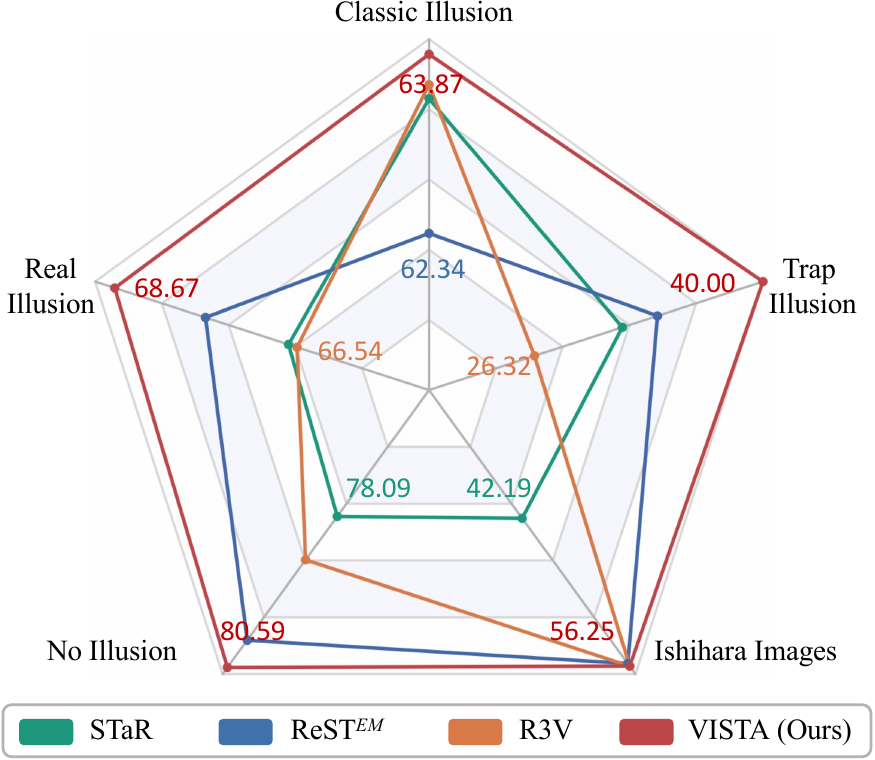}
    \caption{\textbf{Performance comparison between different tuned Qwen2.5-VL-3B models on IllusionBench}. Here, we perform the self-improvement SFT training on the VQA-Rad for one iteration.}
    \label{fig:illusionbench_results}
\end{figure}

\textbf{Model Generalization.}\quad Intuitively, by filtering the hallucinated training data, \texttt{VISTA} can achieve better model generalization. To verify it, we conduct experiments from two aspects: 1) hallucination alleviation and 2) out-of-distribution (OOD) evaluation. First, for 1), we test the models tuned with different self-improvement methods on the popular hallucination evaluation benchmark, IllusionBench~\cite{zhang2025illusionbench}. The benchmark contains 5 categories, including \textit{Classic Illusion}, \textit{Real Scene Illusion}, \textit{No Illusion}, \textit{Ishihara Images}, and \textit{Trap Illusion}. In this experiment, we use the VQA-Rad as the training set, and illustrate the results of tuned Qwen2.5-VL-3B models on IllusionBench in Figure~\ref{fig:illusionbench_results}. Notably, all models are self-trained for one iteration. It can be seen that, while ReST$^{EM}$ performs well in the \textit{No Illusion} category, it suffers from hallucinations in the other categories, indicating that self-training on the hallucination solutions could exacerbate model hallucinations. Conversely, by filtering these undesired solutions, our \texttt{VISTA} outperforms all the other counterparts and achieves the best performance across all hallucination categories. 

\begin{figure}[t]
    \centering
    \includegraphics[width=\linewidth]{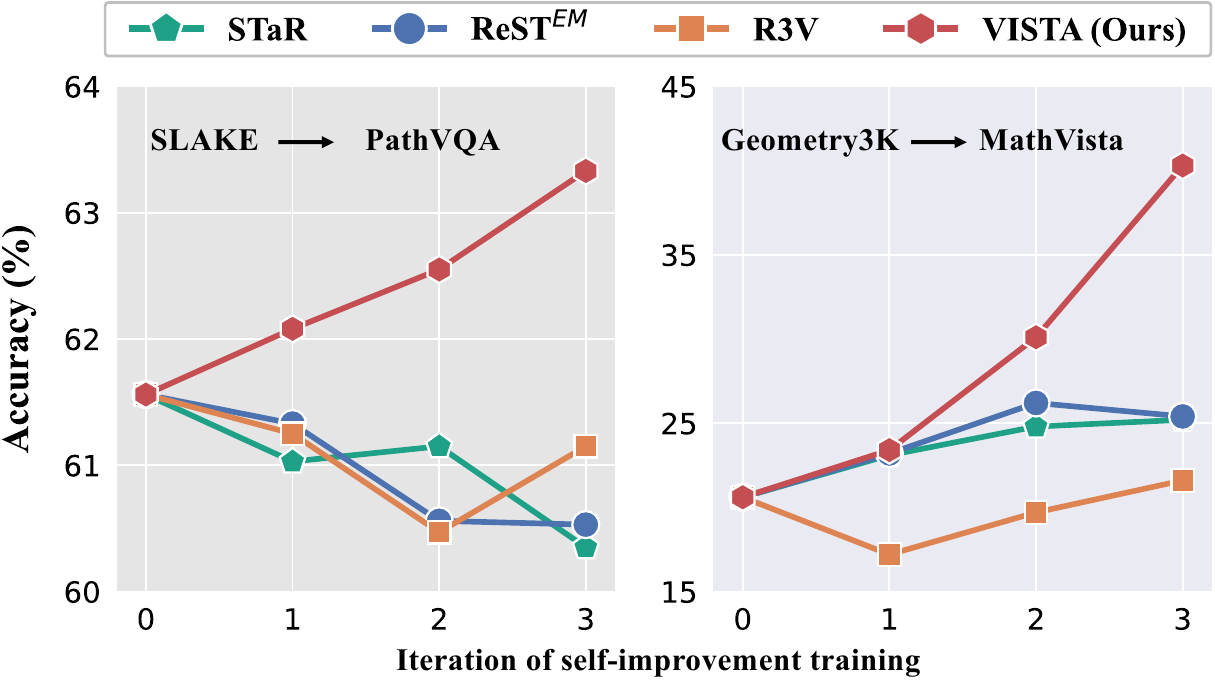}
    \caption{\textbf{Comparison of OOD results} between tuned Qwen2.5-VL-3B models using different self-improvement SFT methods. The x-axis denotes the index of self-improvement iteration.}
    \label{fig:ood_results}
\end{figure}

Second, for 2), we evaluate the models trained on SLAKE on the OOD PathVQA~\cite{he2020pathvqa} test set, while evaluating the models trained on Geometry3K on the MathVista~\cite{lumathvista} test set. The OOD results of different tuned Qwen2.5-VL-3B models are illustrated in Figure~\ref{fig:ood_results}. We can observe that, in medical reasoning tasks, with the self-improvement iteration increasing, the tuned models using all baseline methods generally lead to worse OOD results. This phenomenon is similar to prior work~\cite{wuprogress}. We conjecture that these models may suffer from overfitting due to imbalanced self-training data. Compared to these baselines, our \texttt{VISTA} methods consistently improve the OOD performance across all iterations. 
Similar to medical reasoning, on MathVista, \texttt{VISTA} also yields performance gains of up to \textbf{+19.7\%} over the initial SFT model.

Moreover, to rigorously evaluate the OOD generalization of our method, we assess the tuned Qwen2.5-3B-VL models on several challenging benchmarks: MMMU~\cite{yue2024mmmu}, MathVerse~\cite{zhang2024mathverse}, and BLINK-Twice~\cite{jiang2026blink}. Specifically, for MMMU, we select 5 medicine-related subtasks as the test set. The OOD results after a single round of self-improvement SFT training with different methods are presented in Table~\ref{tab:ood_evaluation}. As shown, \texttt{VISTA-SFT} consistently outperforms all baselines across all OOD settings, demonstrating strong generalizability and validating the ability of VAS to encourage genuine visual grounding.
Overall, these results demonstrate that our \texttt{VISTA} can indeed improve the model generalization effectively.

\begin{table}[t]
\centering
\caption{\textbf{More challenging OOD evaluation} of Qwen2.5-3B-VL after single-round self-improvement SFT with different methods.}
\label{tab:ood_evaluation}
\setlength{\tabcolsep}{9pt}
\resizebox{0.48\textwidth}{!}{%
\begin{tabular}{lcccc}
\toprule
\multirow{2}{*}{\textbf{Method}} & \multicolumn{2}{c}{\textbf{SLAKE$\rightarrow$}} & \multicolumn{2}{c}{\textbf{Geo3K$\rightarrow$}} \\ \cmidrule(lr){2-3} \cmidrule(lr){4-5} 
& MMMU & BLINK & MathVerse & BLINK \\
\midrule
SFT-Seed & 16.34 & 49.35 & 17.08 & 39.17 \\
\midrule
STaR & 16.34 & 49.78 & 16.54 & 39.89 \\
ReST$^\text{EM}$ & 14.04 & 47.62 & 15.52 & 37.95 \\
R3V & 14.78 & 44.73 & 12.34 & 37.81 \\
\rowcolor[RGB]{233,246,255} \bf \texttt{VISTA-SFT} & \textbf{17.64} & \textbf{51.23} & \textbf{21.76} & \textbf{40.26} \\
\bottomrule
\end{tabular}
}
\end{table}

\textbf{\ding{43} Note:} Due to space limitations, we introduce the related works in Appendix~\ref{sec:related_work}, more experimental details in Appendix~\ref{sec:experiment_detail}, and more experiments and analyses in Appendix~\ref{sec:more_experiments}. Please refer to the Appendix for more details.

\section{Conclusion}
\label{sec:conclusion}

In this paper, we reveal that self-improvement training in MLLMs usually suffers from data imbalance and language prior bias problems. To this end, we propose \texttt{VISTA} that effectively harnesses self-improvement in MLLMs via two simple-yet-effective approaches: 1) \textit{prefix resampling} and 2) \textit{vision-aware attention score} (VAS). Specifically, prefix resampling first uses a trajectory recycling strategy to efficiently collect more accurate solutions for difficult queries, while VAS then leverages the internal attention information of MLLMs to quantify and filter the undesired solutions. Extensive experiments show that \texttt{VISTA} not only outperforms the other counterparts by a clear margin across various MLLMs and tasks, \textit{e.g.}, bringing up to \textbf{+13.66\%} average scores for Qwen2.5-VL-3B-Instruct model, but also improves the model generalization effectively.

\section*{Impact Statement}
For the ethics of our work, we take ethical considerations very seriously and strictly adhere to the ICML Ethics Policy. This paper proposes a new vision-aware self-improvement training framework to improve the multimodal reasoning performance of MLLMs. It aims to unleash MLLMs' internal reasoning capabilities, rather than encouraging them to learn privacy knowledge that may cause an ethical problem. Moreover, all base models, training and evaluation datasets used in this paper are publicly available and have been widely adopted by researchers. Thus, we believe that this research will not pose ethical issues. For the potential societal consequences, we believe that there are many potential societal consequences of our work, but none of which we feel must be specifically highlighted here.

\section*{Limitations}
Our work has several potential limitations. First, our prefix resampling strategy retains the early correct prefix and regenerates the subsequent reasoning. While this effectively produces additional correct solutions for challenging queries, it may restrict trajectory diversity when the critical token appears late in the reasoning trajectory, causing the resampled solutions to share substantial overlap with the original. This could potentially lead to overfitting if such cases dominate the training data. Second, due to computational constraints, our experiments are conducted on relatively small-scale models (\textit{e.g.}, 3B and 7B). Validating our \texttt{VISTA} method on larger-scale models (\textit{e.g.}, 30B and 70B) or Mixture-of-Experts (MoE) architectures would further strengthen the credibility and generalizability of our approach, and we leave this exploration to future work.

\section*{Acknowledgements}
We are grateful to the anonymous reviewers and the area chair for their insightful comments and suggestions.
This work was supported in part by the National Key Research and Development Program of China under Grant 2023YFC2705702, in part by the National Natural Science Foundation of China under Grant 62225113, U23B2048 and 625B2132, in part by the Innovative Research Group Project of Hubei Province under Grants 2024AFA017, in part by the Science and Technology Major Project of Hubei Province under Grants 2024BAB046 and 2025BCB026, and in part by the New Cornerstone Science Foundation through the XPLORER PRIZE. This work was also supported by WHU-Kingsoft Joint Lab. Dr Tao’s research is partially supported by NTU RSR and Start Up Grants. The numerical calculations in this paper have been done on the supercomputing system in the Supercomputing Center of Wuhan University.

\nocite{langley00}

\bibliography{icml2026}
\bibliographystyle{icml2026}

\newpage
\appendix
\onecolumn

\section*{Appendix}

\textbf{Roadmap.}\quad In the part, we first introduce the related work in Appendix~\ref{sec:related_work}, and then provide the experimental details in Appendix~\ref{sec:experiment_detail}. Lastly, we present more experiments and analyses in Appendix~\ref{sec:more_experiments}.

\section{Related Work}
\label{sec:related_work}


\textbf{Post-training for Multimodal Reasoning.}\quad Recently, we have witnessed the great success of long-CoT reasoning LLMs, \textit{e.g.}, DeepSeek-R1~\cite{guo2025deepseek} and OpenAI o1~\cite{jaech2024openai}, in a diversity of natural language processing tasks~\cite{shao2024deepseekmath,chen2024huatuogpt,zhong2026achieving}. Motivated by this, extending the advantage of long-CoT reasoning to multimodal context has attracted significant interest~\cite{wang2024exploring,wang2025multimodal,zhu2025internvl3,bai2025qwen2,Qwen3-VL}. To achieve this goal, a common approach is to post-train the MLLMs with explicit reasoning trajectories~\cite{huang2025medvlthinker,xu2025lingshu,feng2025video}. However, such a method highly relies on extensive, high-quality reasoning trajectories, which are usually costly and time-consuming to obtain~\citep{pengregenesis}. While the new emerging Reinforcement Learning with Verifiable Rewards (RLVR) training paradigm~\cite{guo2025deepseek,wen2025reinforcement} does not strictly require the explicit reasoning trajectories, cold-start training with these trajectories can effectively improve the performance and training efficiency~\citep{yangdemystifying}, which also underscores the importance of these trajectories.

\textbf{Self-improvement Training for MLLMs.}\quad To reduce the reliance on explicit reasoning trajectories, a ``self-improvement'' training paradigm has been proposed, where models improve themselves using self-generated data without any external supervision. Prior works mainly focus on the self-improvement of LLMs~\citep{zelikman2022star,yuan2023scaling,huang2023large,gulcehre2023reinforced,wang2024self,hosseiniv,wuprogress,huangself,songmind}. For instance, in the SFT post-training setting, STaR~\citep{zelikman2022star} utilizes few-shot examples to gather self-synthesizing correct reasoning paths for SFT training, while RFT~\citep{yuan2023scaling} and ReST$^{EM}$~\citep{singhbeyond} extend STaR by sampling multiple responses for each question. In the preference learning setting, \citet{pang2024iterative} and \citet{wang2024self} propose to construct preference pairs by using the self-generated correct responses as the pair winners and the incorrect responses as the pair losers. Recent advances attempt to extend self-improvement training to multimodal reasoning~\cite{cheng2025vision,liudiving,guanrstar,deng2025self}. Specifically, beyond ReST$^{EM}$, \citet{cheng2025vision} propose to construct the self-refine and self-select training data for better learning from mistakes.

Despite their effectiveness, we empirically find that these self-improvement methods usually suffer from data imbalance and language prior bias. Some prior studies also recognize these problems and attempt to address them by collecting more correct solutions for difficult queries~\citep{tong2024dart,ding2025mitigating,koh2025adastar} and designing metrics to measure the language priors~\citep{he2025cracking,liu2025more}. However, these efforts either fail to exploit prior failed solutions fully or rely on external models and additional computational overhead to estimate the language prior bias. Different from them, we propose two simple-yet-effective approaches to address these problems efficiently. On the one hand, instead of solely allocating more trials to difficult queries, our \textit{prefix resampling} strategy makes full use of the partial correct prefix context to achieve efficient data collection. On the other hand, although simple, our proposed \textit{Vision-Aware Score} can accurately quantify the model’s attention to the visual context within seconds.

Notably, our work differs from the concurrent work~\cite{zhonghsir} in three key aspects: 1) \textit{Different research objectives}. The concurrent work focuses on self-improvement in LLMs and mainly addresses the overthinking issue, whereas we study self-improvement in MLLMs and target visual hallucinations specific to multimodal reasoning. 2) \textit{Different methodologies}. Although both works involve alleviating data imbalance by reusing prior failed solutions, the concurrent work relies on ground-truth answers to verify intermediate reasoning steps, while our method does not require ground-truth supervision and instead leverages the model’s self-calibration to identify critical tokens. 3) \textit{Different experimental settings}. The concurrent work evaluates on text-only tasks, whereas we conduct experiments on multimodal VQA tasks.

 \begin{figure}[t]
     \centering
     \includegraphics[width=1\linewidth]{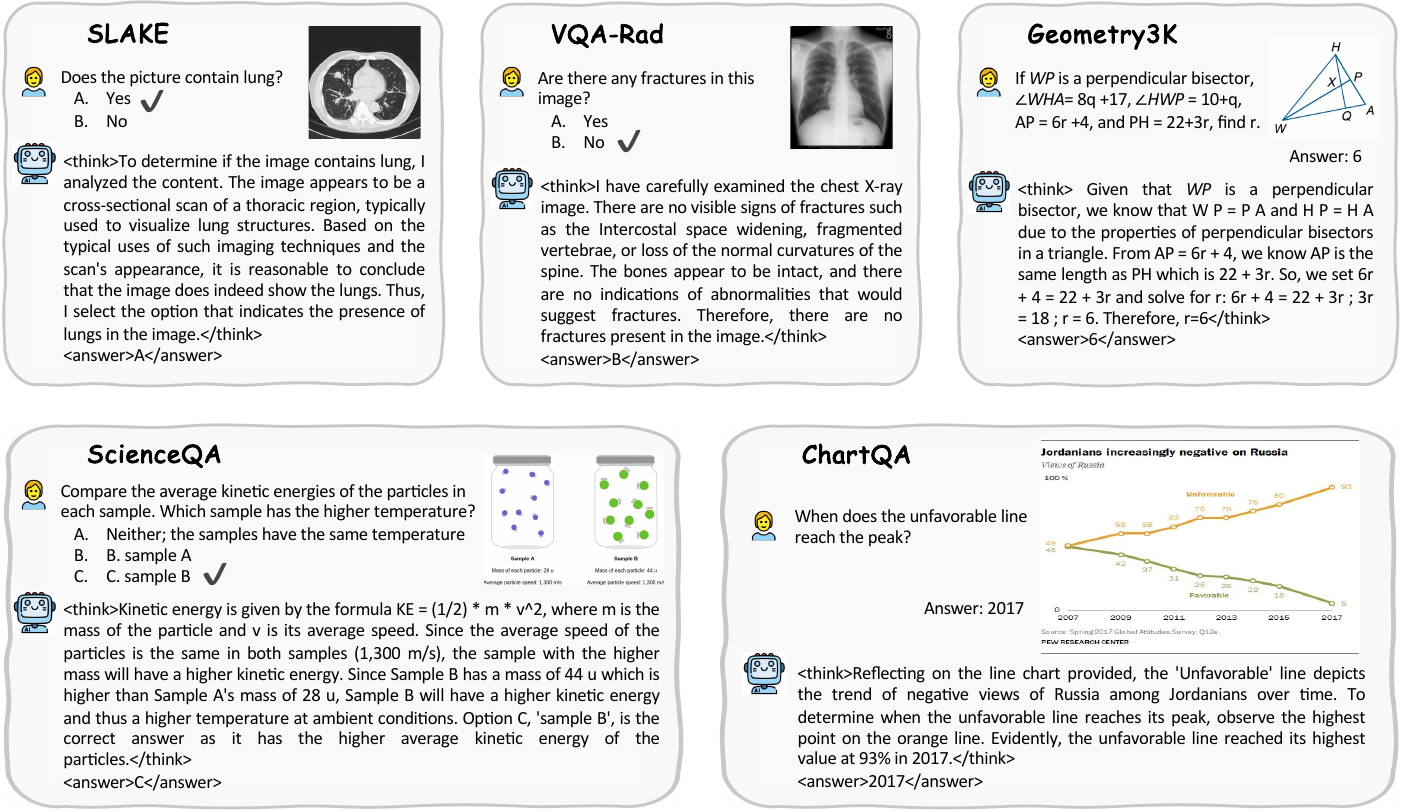}
     \caption{Examples of self-generated long-CoT data in various multimodal reasoning tasks.}
     \label{fig:training_data_case}
 \end{figure}

\section{More Experimental Details}
\label{sec:experiment_detail}

\subsection{Dataset Details} 
\label{sec:dataset_details}
In this work, we conduct experiments on a diverse range of multimodal reasoning benchmarks. Here, we first provide some self-generated long-CoT examples in Figure~\ref{fig:training_data_case}, and then introduce the descriptions of these tasks as follows:

\begin{itemize}[leftmargin=10pt, itemindent=0pt]
	\item \textbf{SLAKE}: SLAKE~\citep{liu2021slake} is a widely-used medical visual question-answering (VQA) task, which contains both bilingual closed-ended and open-ended questions across various modalities and human body parts. In this work, we only use the closed-ended English questions and their corresponding answers. By filtering the data, we finally obtain a training set with 1,681 samples, and a test set with 355 samples.
	\item \textbf{VQA-Rad}: VQA-Rad~\citep{lau2018dataset} is also a popular multimodal medical reasoning task, which contains 2,248 question-answer pairs and 315 radiology images. The dataset includes both open-ended questions and binary ``yes/no'' questions. Similar to SLAKE, we also use the closed-ended questions for model training and evaluation. Specifically, the final training and test sets have 940 and 251 samples, respectively.
	\item \textbf{Geometry3K}: Geometry3K~\citep{lu2021inter} comprises 3,002 geometry problems. Here, to verify the effectiveness of our \texttt{VISTA} in more challenging tasks, we use a variant of Geometry3K\footnote{https://huggingface.co/datasets/hiyouga/geometry3k} that converts the multiple-choice problems into open-ended ones. Specifically, there are 2,101 queries in the training set, and 601 queries in the test set.
	\item \textbf{ScienceQA}: ScienceQA~\citep{lu2022learn} is a large-scale science QA task, which contains 21,208 multimodal multiple-choice questions from elementary and high school science curricula. Due to the limited computational resources, we use a subset of ScienceQA for model training and evaluation. Specifically, only the closed-ended questions with image context and grades higher than 6 are used in this work, resulting in 2,635 training samples and 813 test samples.
	\item \textbf{ChartQA}: ChartQA~\citep{masry2022chartqa} is a large-scale chart reasoning benchmark covering 9.6K human-written questions and 23.1K generated questions. This benchmark evaluates MLLMs' visual and logical reasoning capabilities. Similar to ScienceQA, we also sample a subset of ChartQA for a faster experimental validation. In particular, only the human-written questions with digit answers are selected, resulting in 3,233 training samples and 1,026 test samples. 
	\item \textbf{PathVQA}: PathVQA~\citep{he2020pathvqa} is a challenging medical VQA benchmark, containing a total of 5K pathology images and 32K question-answer pairs. We use the 3,362 closed-ended questions from the original test set to evaluate the OOD performance of models trained on SLAKE.
	\item \textbf{MathVista}: MathVista~\citep{lumathvista} is a consolidated mathematical reasoning benchmark, which consists of three newly created datasets, 9 MathQA datasets, and 19 VQA datasets from the literature. In total, MathVista includes 6K examples collected from 31 different datasets. We use the 1,000 closed-ended test samples to evaluate the OOD performance of models trained on Geometry3K.
	\item \textbf{IllusionBench}: IllusionBench~\citep{zhang2025illusionbench} is a large-scale visual illusion benchmark, encompassing over 1K images, 5K question-answer pairs, and 1K golden text descriptions, covering the presence, causes, and content of illusions. Specifically, it contains five categories: 1) \textit{Classic Cognitive Illusion}, including blur, distortion, paradox, and fictitious illusions—key examples of traditional synthetic illusions. 2) \textit{Real Scene Illusion}, containing images that depict real-world objects and scenes, with unique and definite semantic descriptions. 3) \textit{No Illusion}, depicting diverse subjects such as people, landscapes, and objects. 4) \textit{Ishihara Images}, verified by vision-healthy individuals, where the patterns convey unique and definite semantics. 5) \textit{Trap Illusion}, edited versions of classic visual illusions, resembling them in appearance but differing in physical properties.
\end{itemize}

Notably, for each task, the reasoning template is similar to that in DeepSeek-R1, \textit{i.e.}, the reasoning process and answer are enclosed with <think></think> and <answer></answer> tags. Specifically, we use the following system prompt in our work:

\begin{tcolorbox}
[colback=lightgray!20,colframe=darkgray!80,title= System Prompt]
\label{tab:system_prompt}
A conversation between User and Assistant. The user asks a question, and the Assistant solves it. The assistant first thinks about the reasoning process in the mind and then provides the user a concise final answer in a short word or phrase. The reasoning process and answer are enclosed within <think> </think> and <answer> </answer> tags, respectively, i.e., <think> reasoning process here </think><answer> answer here </answer>.
\end{tcolorbox}

\subsection{Training and Evaluation Details} 
\label{sec:training_details}

During the SFT training, we fine-tune each model with a batch size of 32 and a peak learning rate of 1e-5. The total training epoch is set to 3. The max image pixels are set to $512 \times 512$. In the DPO training phase, the batch size is set to 16, and the peak learning rate is set to 1e-5. All models are trained for 3 epochs. Both SFT and DPO training are performed using the popular \texttt{LLaMA-Factory}\footnote{https://github.com/hiyouga/LLaMA-Factory} toolkit, following prior work~\cite{zhong2025kaft}. As for GRPO training, the batch size is set to 32, and the rollout size for each query is set to 10. The peak learning rate is set to 1e-6. The maximum output length during on-policy sampling is set to 512. Each model is trained for 3 epochs. We use the \texttt{EasyR1}\footnote{https://github.com/hiyouga/EasyR1} as the training framework of GRPO. Notably, for the model optimizer of all settings, we keep the vision encoder and multimodal projector fixed, and only update the parameters of the LLM backbone. All experiments are conducted on 8 NVIDIA A800 (80GB) GPUs.

For the model evaluation, we use the greedy decoding method, \textit{i.e.}, temperature set to 0 for reproducibility. The maximum output length is set to 2,048. All models are evaluated in a zero-shot manner. We extract the final answer enclosed within \texttt{<answer></answer>}. If no valid answer is extracted, the response is considered incorrect.

\section{More Experiments and Analyses}
\label{sec:more_experiments}

 \begin{figure}[t]
     \centering
     \includegraphics[width=1\linewidth]{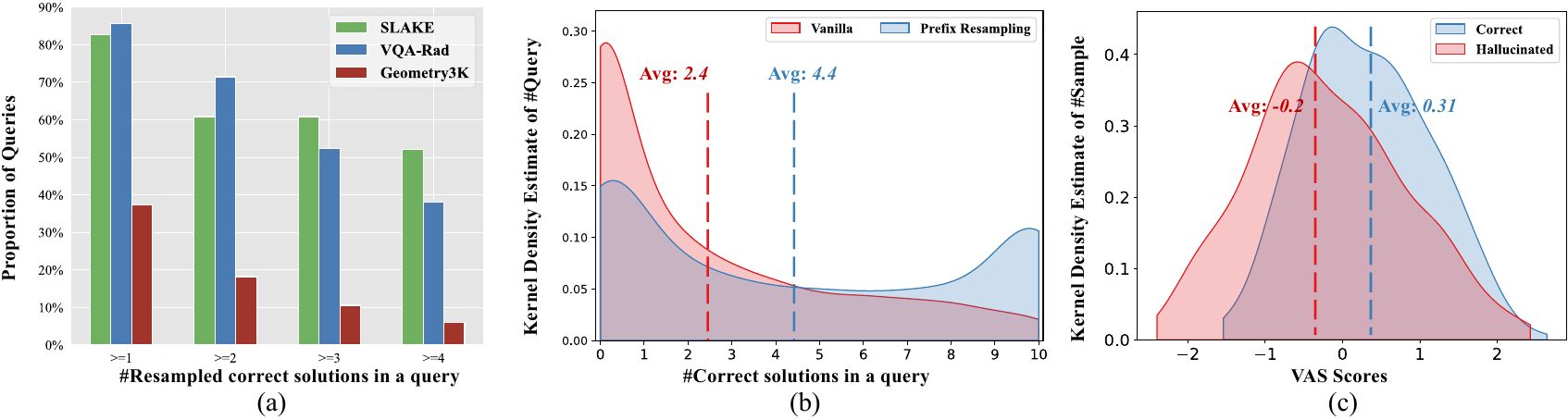}
     \caption{\textbf{(a)} Sampling success rate of our prefix resampling on the hardest queries (\textit{i.e.}, without any prior correct solutions). The x-axis denotes the number of resampled correct solutions in a query. \textbf{(b)} Distribution of the number of correct tokens in a query before and after using our prefix resampling strategy. Here, we show the results on the challenging Geometry3K task. \textbf{(c)} Distribution of our VAS scores in correct and hallucinated solutions from the VQA-Rad task. Notably, Qwen2.5-VL-3B-Instruct is used in these experiments.}
     \label{fig:method_analysis}
 \end{figure}

\subsection{Analysis of {Prefix Resampling} Strategy}
\label{sec:resampling_analysis}
\textbf{Effectiveness of prefix resampling.}\quad
To verify whether our \textit{prefix resampling} can alleviate the data imbalance problem, we calculate the sampling success rate on the hardest queries, which did not obtain any correct solutions during the $K$-times self-generation processes. Figure~\ref{fig:method_analysis} (\textbf{a}) illustrates the results of Qwen2.5-VL-3B-Instruct on several benchmarks, where the x-axis denotes the number of resampled correct solutions in a query, and the y-axis denotes the proportion of queries. As seen, although the vanilla data collection method struggles to sample any correct solutions for these difficult queries, our prefix resampling strategy can effectively improve the sampling success rate. Specifically, for both SLAKE and VQA-Rad tasks, more than 80\% of these difficult queries can resample at least one correct solution, and nearly 40\% of these queries have 4 or more correct solutions. While for the harder Geometry3K task, relatively fewer correct solutions are successfully resampled, our prefix resampling strategy still collects at least one correct solution for 37\% difficult queries. To have a close look, taking the challenging Geometry3K task as an example, we further compare the correct solution distribution before and after using the prefix resampling strategy. As illustrated in Figure~\ref{fig:method_analysis} (\textbf{b}), with the help of prefix resampling, we can collect more correct solutions for difficult queries, \textit{i.e.}, improving the average number of correct solutions per query from 2.4 to 4.4. These results demonstrate that our prefix resampling indeed alleviates the data imbalance effectively.

\textbf{Effectiveness of the token position swap mechanism.}\quad To obtain the rewritten solution, we perform a swap operation on the original negative solution. A natural question is whether this swap strategy is effective and why alternative perturbation strategies, such as masking or noise injection, are not adopted instead. We clarify that the swap operation leverages the model's internal self-calibration in response to context changes, rather than introducing random perturbations. Specifically, by swapping the visual and instruction token order, the context structure is altered, exposing model uncertainty about the critical token~\cite{lincritical}. In contrast, masking and noise injection may corrupt valid content and cause significant semantic shifts, thereby impairing the model's precision in locating critical tokens. To further validate the effectiveness of the swap method, we replace it with masking and noise injection in \texttt{VISTA} and conduct experiments on Qwen2.5-3B-VL. The comparative results under single-round self-improvement SFT training are presented in Table~\ref{tab:swap_comparison}. As shown, the swap method achieves superior performance compared to both masking and noise injection. Additionally, it is worth noting that although masking and noise injection perform slightly worse, they still outperform the baseline ReST$^\text{EM}$, suggesting that \texttt{VISTA} exhibits a reasonable degree of robustness to noise in critical token detection.

\begin{table}[t]
\centering
\begin{minipage}[t]{0.5\textwidth}
\centering
\caption{\textbf{Perturbation strategy comparison}. Results on Qwen2.5-3B-VL after single-round self-improvement SFT training.}
\label{tab:swap_comparison}
\setlength{\tabcolsep}{14.8pt}
\resizebox{0.8\textwidth}{!}{%
\begin{tabular}{lcc}
\toprule
\textbf{Method} & \textbf{SLAKE} & \textbf{VQA-Rad} \\
\midrule
ReST$^\text{EM}$ & 76.06 & 69.72 \\
\midrule
\rowcolor[RGB]{233,246,255} \textbf{\texttt{VISTA}-Swap} & \textbf{80.85} & \textbf{74.10} \\
\texttt{VISTA}-Mask & 78.31 & 70.12 \\
\texttt{VISTA}-Noise & 78.87 & 70.52 \\
\bottomrule
\end{tabular}
}
\end{minipage}
\hfill
\begin{minipage}[t]{0.45\textwidth}
\centering
\caption{\textbf{Effect of the Top-$k$ setting}. Results on Qwen2.5-3B-VL after single-round self-improvement SFT training.}
\label{tab:topk_effect}
\setlength{\tabcolsep}{14pt}
\resizebox{0.8\textwidth}{!}{%
\begin{tabular}{ccc}
\toprule
\textbf{Top-$k$} & \textbf{SLAKE} & \textbf{VQA-Rad} \\
\midrule
1 & 80.28 & 72.91 \\
\rowcolor[RGB]{233,246,255} \textbf{5} & \textbf{80.85} & \textbf{74.10} \\
10 & 80.00 & 71.71 \\
50 & 78.03 & 69.72 \\
\bottomrule
\end{tabular}
}
\end{minipage}
\end{table}

\textbf{Effect of the Top-$k$ setting.}\quad During prefix resampling, we use the Top-5 token predictions to detect critical tokens. Here, we investigate the effect of the top-$k$ setting by varying $k \in \{1, 5, 10, 50\}$ and comparing the resulting performance of \texttt{VISTA}. The comparative results on Qwen2.5-3B-VL are presented in Table~\ref{tab:topk_effect}. We observe that when $k$ is too small ($k{=}1$), the model suffers from over-calibration, truncating at very early prefix tokens and degrading sampling efficiency. Conversely, when $k$ is too large ($k{=}50$), the model overlooks many critical tokens, reducing the effectiveness of prefix resampling. With $k{=}5$, the model achieves the best overall performance, and we therefore adopt it as our default setting.

  \begin{figure}[t]
     \centering
     \includegraphics[width=\linewidth]{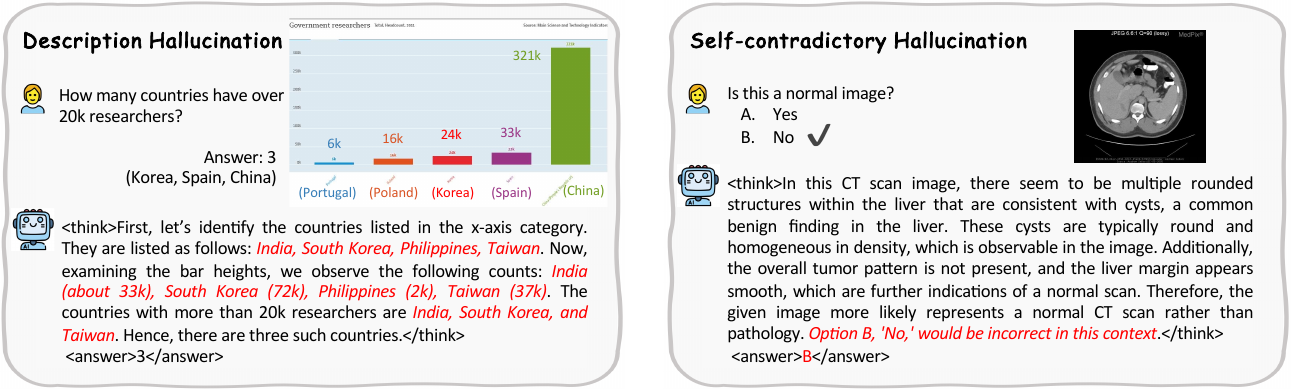}
     \caption{Examples of two types of self-generated hallucinated solutions, where the hallucinated contents are highlighted in red.}
     \label{fig:hallucinated_data_case}
 \end{figure}

 \begin{figure}[t]
     \centering
     \includegraphics[width=0.85\linewidth]{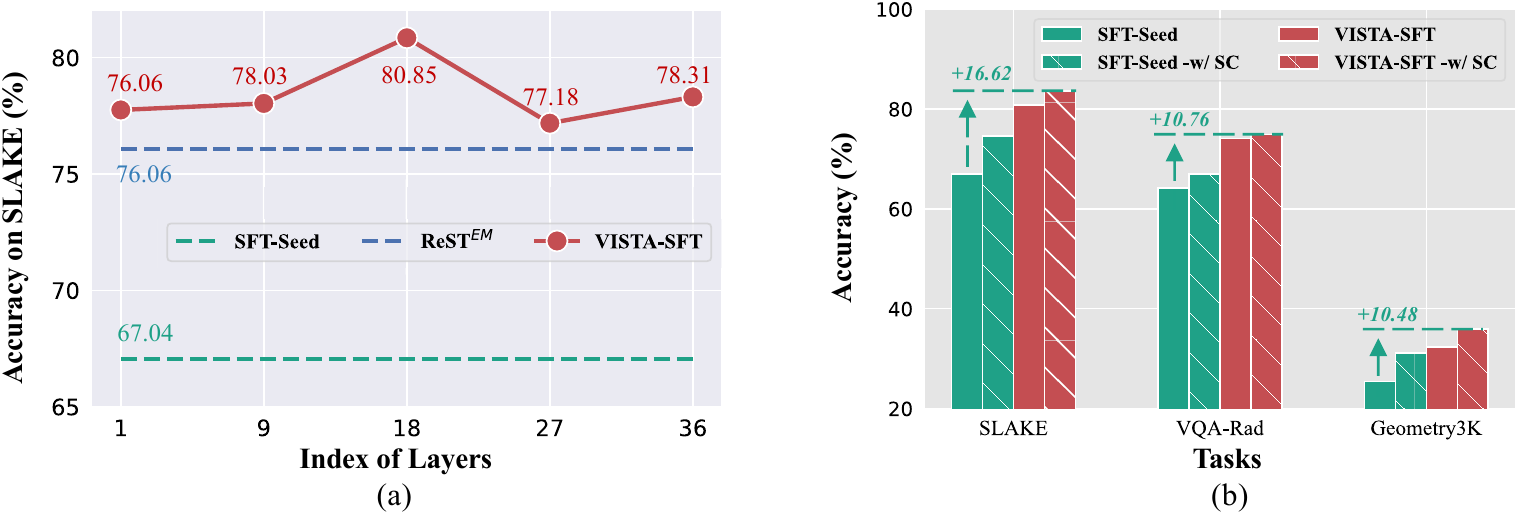}
     \caption{\textbf{(a)} Analysis of layer depth for calculating VAS scores. \textbf{(b)} Performance comparison between with and without the self-consistency method. Notably, we report the results of Qwen2.5-VL-3B-Instruct models self-trained for one iteration.}
     \label{fig:more_analysis}
 \end{figure}

\subsection{Analysis of {VAS} Metric}
\label{sec:vas_analysis}

\textbf{Reliability of VAS.}\quad By analyzing the quality of self-generated candidate solutions, we found that some solutions suffer from visual hallucinations~\cite{huang2024visual,bai2024hallucination,zheng2024thinking}, even though they produce the correct answers. As shown in examples of Figure~\ref{fig:hallucinated_data_case}, during the reasoning process, MLLMs may describe some things that do not exist in the image (denoted as ``Description Hallucination''), or exhibit a conflict between the reasoning chain and its final answer (denoted as ``Self-contradictory Hallucination''). We speculate that the main reason for this problem may lie in MLLMs' strong language priors, which affect the focus and usage of visual cues. To this end, we propose a vision-aware attention score (VAS) to quantify it. Here, to verify whether our metric can identify the hallucinated solutions, we first prompt the GPT-4o to determine whether the candidate solution is hallucinated. In practice, we randomly sample 500 queries from the training set of VQA-Rad, which are then used to collect the candidate solutions by Qwen2.5-VL-3B-Instruct. For each self-generated candidate solution of these queries, we enforce GPT-4o to detect whether there are any non-existent contents in the reasoning traces or whether the reasoning chain conflicts with the final answer. By doing so, we finally collected 586 hallucinated solutions. Correspondingly, 586 non-hallucinated solutions are randomly selected for comparison.
For these paired data, we calculate their VAS scores and illustrate the distribution in Figure~\ref{fig:method_analysis} (\textbf{c}). Obviously, the hallucinated solutions have a lower VAS score than the correct ones. Specifically, the average VAS score of hallucinated solutions is -0.2, while the average VAS score of correct solutions is 0.31. That is, our VAS score can effectively distinguish between hallucinated solutions and correct ones, proving its reliability.

\textbf{Impact of layer depth for calculating VAS scores.}\quad As mentioned in \S\ref{sec:method}, we use the attention output of the middle layer of MLLMs to calculate the VAS scores. Here, we investigate the impact of different layer depths by comparing the performance of tuned Qwen2.5-VL-3B models using different \texttt{VISTA} configurations on SLAKE. Specifically, for the Qwen2.5-VL-3B-Instruct containing 36 layers, we vary the layer used for calculating VAS scores across \{1, 9, 18, 27, 36\} and illustrate the comparative results in Figure~\ref{fig:more_analysis} (\textbf{a}). For reference, we also include the results of SFT-Seed and ReST$^{EM}$ methods. All models are self-improved for one iteration. As seen, \texttt{VISTA} with varied layer depth can consistently outperform the other baseline methods, indicating that \texttt{VISTA} is relatively robust to the choice of layer. Moreover, when using the middle layer (\textit{i.e.}, 18-th layer), \texttt{VISTA} achieves the best performance. 
We conjecture that the middle layer encodes richer and more useful semantic information~\citep {skeanlayer,azaria2023internal,liu2019linguistic}, and plays a crucial role in processing visual information~\cite{jiang2025devils}, thus resulting in more accurate VAS scores. Based on these observations, we choose to adopt the middle layer of $\mathcal{M}_{t-1}$ for calculating VAS scores in this work.

\subsection{Comparison and Compatibility with Inference-time Method}
\label{sec:self_consistency}

Since the goal of our work is to propose a self-improvement training, rather than to optimize inference, we do not compare our \texttt{VISTA} with inference-time methods, such as Self-Consistency (SC)~\citep{wangself}, in the main experiments. Nevertheless, considering that SC is widely used to enhance models' reasoning performance, we include it in this experiment. Specifically, for the implementation of SC, we sample five solutions for each test query and select the majority-vote answer as the final prediction. Figure~\ref{fig:more_analysis} (\textbf{b}) presents the comparative results of tuned Qwen2.5-VL-3B models on several benchmarks, where ``-w/ SC'' means using the SC method. As seen, by increasing the test-time compute, SC indeed improves the multimodal reasoning performance of SFT-Seed models effectively. However, it still underperforms our zero-shot \texttt{VISTA} method. More encouragingly, combining the \texttt{VISTA} and SC methods consistently yields further performance improvements. For instance, in the SLAKE task, \texttt{VISTA} equipped with SC achieves a +16.62\% performance gain over the SFT-Seed model. These results prove the compatibility of our \texttt{VISTA} with the inference-time SC method. 

\begin{wraptable}{r}{0.5\textwidth}
\centering
\vspace{-8pt}
\caption{\textbf{Wall-clock time comparison} (in minutes) of different self-improvement SFT methods on SLAKE using Qwen2.5-3B-VL.}
\label{tab:computational_cost}
\setlength{\tabcolsep}{10pt}
\resizebox{0.48\textwidth}{!}{%
\begin{tabular}{lccc}
\toprule
\textbf{Method} & \textbf{Data Collection} & \textbf{Training} & \textbf{Total} \\
\midrule
STaR & 1.3 & 2.5 & 3.9 \\
ReST$^\text{EM}$ & 2.3 & 26.2 & 28.6 \\
R3V & 3.3 & 49.7 & 53.0 \\
\midrule
\rowcolor[RGB]{233,246,255} \textbf{\texttt{VISTA-SFT}} & 12.0 & 17.2 & 29.2 \\
\bottomrule
\end{tabular}
}
\end{wraptable}

\subsection{Efficiency of \texttt{VISTA}}
\label{sec:efficiency}
A potential concern regarding our \texttt{VISTA} framework is its efficiency, as it requires additional forward passes of MLLMs. We acknowledge that \texttt{VISTA} introduces some computational overhead during data collection. However, it is worth noting that \texttt{VISTA} simultaneously reduces the training data volume, thereby improving training efficiency. Taking Qwen2.5-3B-VL on SLAKE as an example, we report the wall-clock time (in minutes) of different self-improvement SFT methods in Table~\ref{tab:computational_cost}. As shown, the total cost of \texttt{VISTA} is comparable to that of ReST$^\text{EM}$ and significantly lower than that of R3V, while achieving substantially better performance. We therefore consider the additional inference overhead to be a worthwhile trade-off given the consistent performance gains.

  \begin{figure}[h]
     \centering
     \includegraphics[width=1\linewidth]{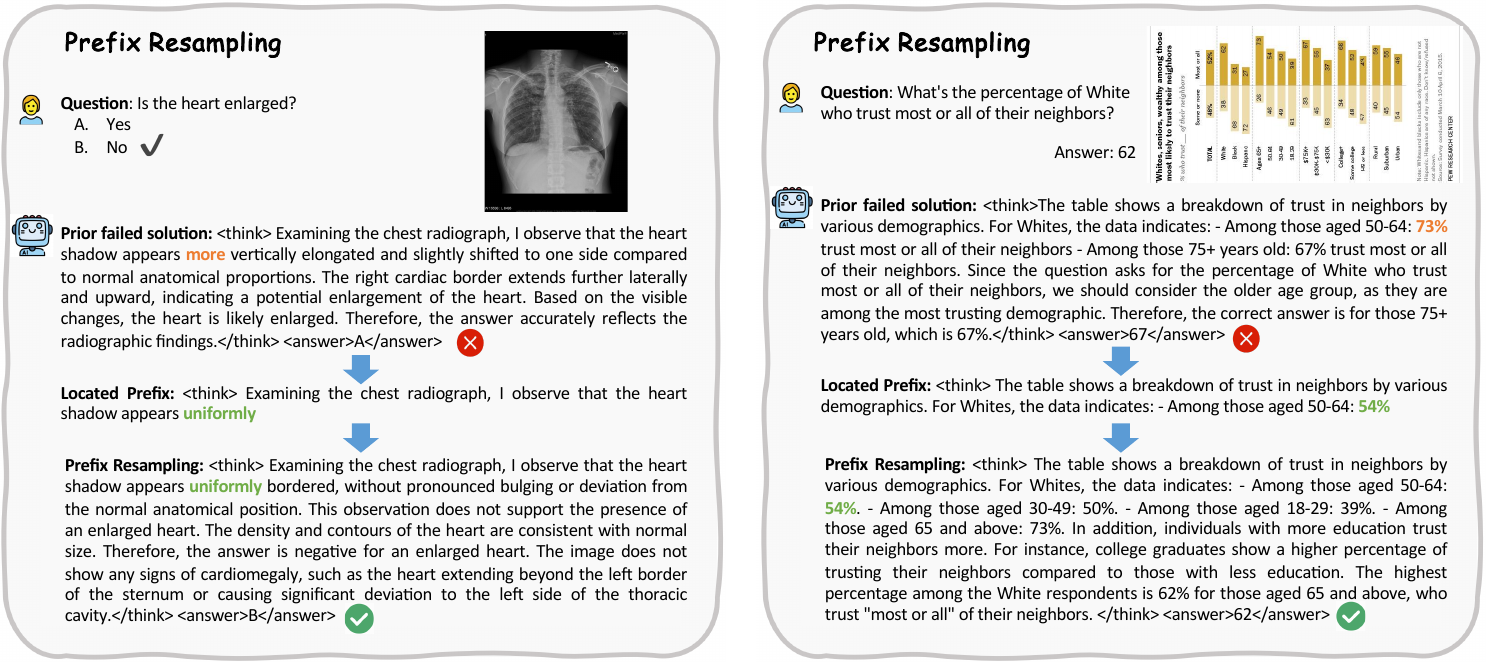}
     \caption{Illustration of our prefix resampling strategy. Notably, we use examples from the SLAKE (Left) and ChartQA (Right).}
     \label{fig:prefix_resampling_case}
 \end{figure}

 \subsection{Case Study}
 \label{sec:case_study}
 
\begin{figure}[H]
     \centering
     \includegraphics[width=1\linewidth]{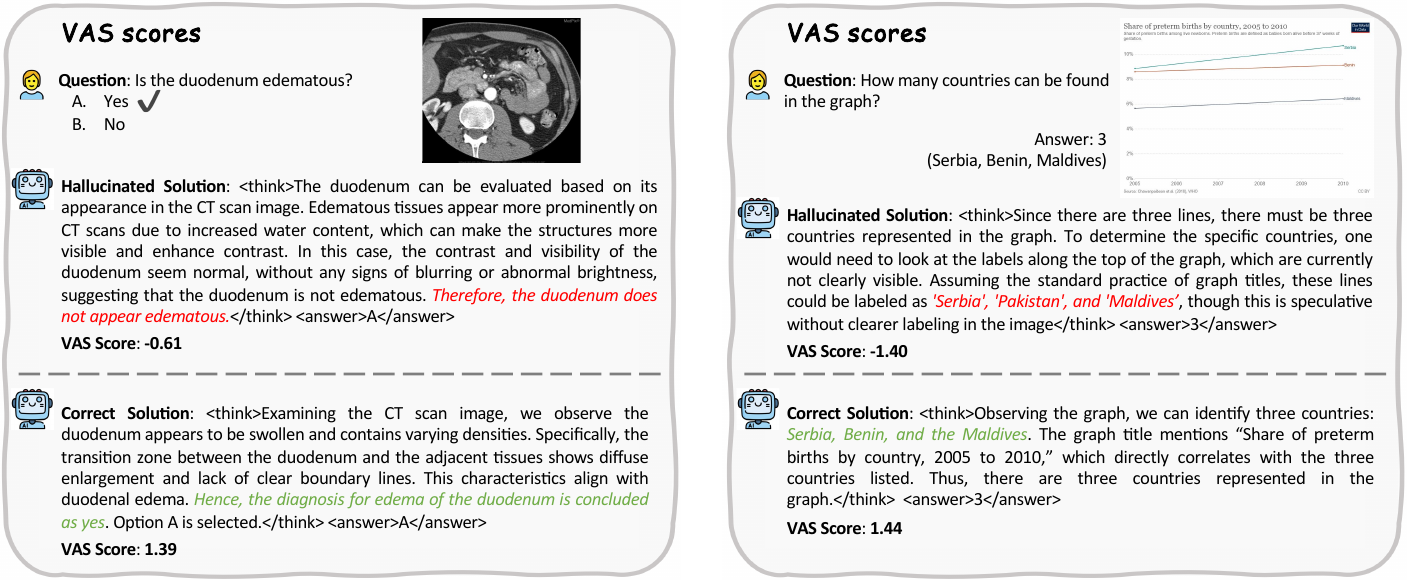}
     \caption{Examples of self-generated solutions with our VAS scores in the VQA-Rad (Left) and ChartQA (Right) tasks.}
     \label{fig:vas_score_case}
 \end{figure}

\begin{figure}[H]
     \centering
     \includegraphics[width=1\linewidth]{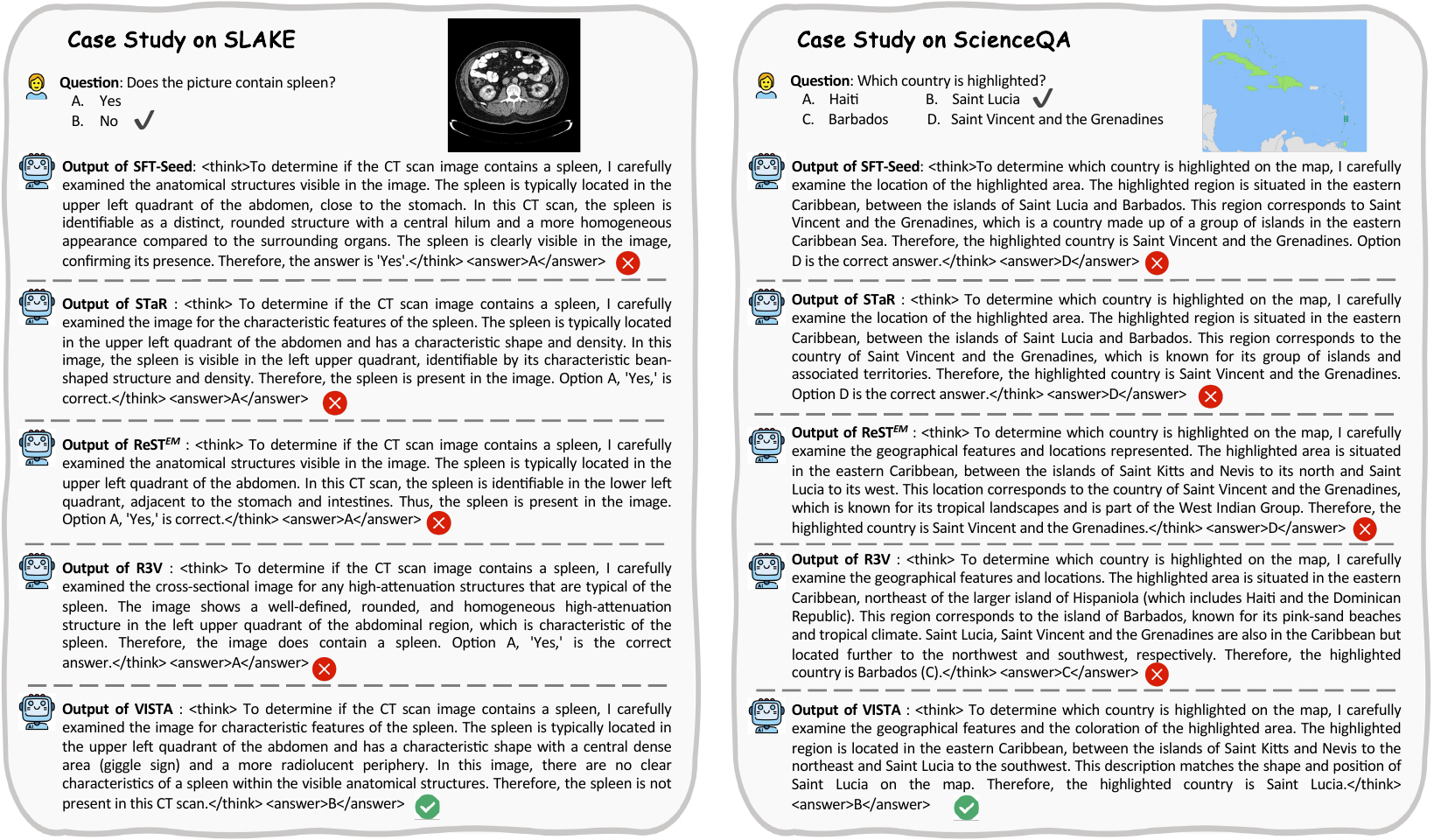}
     \caption{Examples of solutions predicted by Qwen2.5-VL-3B models tuned with different self-improvement training methods.}
     \label{fig:case_study}
 \end{figure}

 To better understand our methods and demonstrate their effectiveness, we provide some cases in this part. First, Figure~\ref{fig:prefix_resampling_case} shows the comparison of self-generated solutions between before and after using our prefix sampling strategy. We present the self-generated results of Qwen2.5-VL-3B-Instruct on the SLAKE and ChartQA tasks. As seen, our prefix sampling can accurately locate the critical tokens and salvage valid initial reasoning from failed attempts by continuing sampling from the correct prefix contexts. Figure~\ref{fig:vas_score_case} shows the comparison of VAS scores between correct and hallucinated candidate solutions in the VQA-Rad and ChartQA tasks. Moreover, we also compare the test results of models trained using different methods and illustrate the comparison in Figure~\ref{fig:case_study}. Obviously, with the help of \texttt{VISTA}, Qwen2.5-VL-3B models can produce more reliable reasoning traces and make more accurate predictions. These qualitative analyses also prove \texttt{VISTA}'s effectiveness.

  \begin{algorithm}[t]
   \footnotesize
 	\renewcommand{\algorithmicrequire}{\textbf{Input:}}
 	\renewcommand{\algorithmicensure}{\textbf{Output:}}
 	\caption{Self-improvement Training with \texttt{VISTA}}
 	\begin{algorithmic}[1]
             \STATE \textbf{Input:} base model $\mathcal{M}_{base}$, training dataset $\mathcal{D}=\{(x_i, y_i)\}$, where $x_i = \{x^{\text{sys}}_i, x^{\text{vis}}_i, x^{\text{ins}}_i\}$
             \STATE \textbf{Output:} self-improved model $\mathcal{M}_T$
        \STATE Obtain a seed reasoning dataset by prompting $\mathcal{M}_{base}$ to generate correct solutions for $\mathcal{D}$ via some few-shot examples
 		\STATE Fine-tune $\mathcal{M}_{base}$ on the seed dataset to get initial reasoning model $\mathcal{M}_0$
 		\FOR{$t \in [1,T]$}
             \STATE \textcolor{lightgray}{\# Data Collection}
 			\STATE Obtain $K$ candidate solutions $\{(r^k_i, \hat{y}^k_i)\}^K_{k=1}$ generated by $\mathcal{M}_{t-1}$ for each $x_i \in \mathcal{D}$
             \STATE Verify the correctness of candidate solutions, and split them into two sets: \\
             \quad positive set $\mathcal{D}^{p}_t=\{(x_i, r^{k_p}_i, \hat{y}^{k_p}_i) \,|\, x_i \in \mathcal{D}; k_p \in [1,\,K]; \mathbb{I}(\hat{y}^{k_p}_i, y_i)=1\}$ \\
             \quad negative set $\mathcal{D}^{n}_t=\{(x_i, r^{k_n}_i, \hat{y}^{k_n}_i) \,|\, x_i \in \mathcal{D}; k_n \in [1,\,K]; \mathbb{I}(\hat{y}^{k_n}_i, y_i)=0\}$ 
         \STATE
         \STATE \colorbox[RGB]{233,246,255}{\textcolor{gray}{\# \textit{Prefix Resampling}}}
         \FOR{Each sample $(x_i, r^{k_n}_i, \hat{y}^{k_n}_i) \in \mathcal{D}^{n}_t$}
          \STATE Swap the order of visual tokens, \textit{e.g.}, from ``$x^{\text{vis}}_i + x^{\text{ins}}_i$'' to ``$x^{\text{ins}}_i + x^{\text{vis}}_i$''
          \STATE Feed the paraphrased sample into $\mathcal{M}_{t-1}$ to obtain the Top-5 token predictions for each token $o_n$ in $r^{k_n}_i$
           \FOR{Each token $o_n \in r^{k_n}_i$}
     		\IF{$o_n \notin \text{Top}_5(o_{n-1})$} 
                 \STATE Replace the $o_n$ with the new Top-1 predicted token $o^{'}_n=\text{Top}_1(o_{n-1})$ and truncate the subsequent reasoning steps
                 \STATE Concatenate the original query with the truncated reasoning traces as a prefix
                 \STATE \textbf{Break}
             \ELSE
                 \STATE \textbf{Continue}
             \ENDIF
         \ENDFOR
         \STATE Feed the prefix context into $\mathcal{M}_{t-1}$ to resample $J$ solutions and add the correct solutions into $\mathcal{D}^{p}_t$
         \ENDFOR
         \STATE
         \STATE \colorbox[RGB]{233,246,255}{\textcolor{gray}{\# \textit{Vision-aware Attention Score}}}
         \FOR{Each query $x_i \in \mathcal{D}^{p}_t$}
             \STATE Calculate the \textit{VAS} score $\text{\textbf{VAS}}^k_i$ for each correct solution $(r^{k_p}_i, \hat{y}^{k_p}_i)$ as Eq.~\ref{eq:vas}
             \STATE Update the dataset $\mathcal{D}^{p}_t$ by filtering out the undesired solution with $\text{\textbf{VAS}}^k_i < \tau$
         \ENDFOR        
         \STATE
         \STATE \textcolor{lightgray}{\# SFT Training}
         \STATE Fine-tune $\mathcal{M}_{base}$ with $\mathcal{L}_\text{SFT}$ in Eq.~\ref{eq:sft_loss} on the $\mathcal{D}^{p}_t$
         \STATE \textcolor{lightgray}{\# or DPO Training}
         \STATE Obtain a pairwise dataset $\mathcal{D}^{\text{pairs}}_t=\{(x_i, r^{k_p}_i, \hat{y}^{k_p}_i, r^{k_n}_i, \hat{y}^{k_n}_i)\,|\, k_p, k_n \in [1,\,K)\}$, where $(r^{k_p}_i, \hat{y}^{k_p}_i) \sim \mathcal{D}^{p}_t$ and $(r^{k_n}_i, \hat{y}^{k_n}_i) \sim \mathcal{D}^{n}_t$
         \STATE Train $\mathcal{M}_{base}$ with $\mathcal{L}_\text{DPO+NLL}$ in Eq.~\ref{eq:dpo_loss} on $\mathcal{D}^{\text{pairs}}_t$
     \ENDFOR
 	\end{algorithmic}
 \label{alg:vista}
 \end{algorithm}


\end{document}